\documentclass[USenglish]{article}

\usepackage[utf8]{inputenc}
\usepackage[small]{dgruyter}
\usepackage{microtype}
\usepackage{linguex}
\usepackage[authoryear]{natbib}
\usepackage{booktabs}
\setcitestyle{notesep={; },round,aysep={},yysep={;}}

\renewcommand*{\thefootnote}{\fnsymbol{footnote}}
\begin{document}
  \articletype{Article}

  \author[1]{Amir Zeldes}
  \author[2]{Jessica Lin}
  \runningauthor{Zeldes \& Lin}
  \affil[1]{Georgetown University, \texttt{amir.zeldes@georgetown.edu}
  }
  \affil[2]{Georgetown University, \texttt{yl1290@georgetown.edu}
  }
  
  \title{What makes an entity salient in discourse?$^*$}
  \runningtitle{What makes an entity salient in discourse?}
  \abstract{Entities in discourse vary in salience: main participants, objects and locations stay prominent, while others are quickly forgotten, raising questions about how humans signal and infer discourse-level salience. Using a graded operationalization of discourse-level salience based on summary-worthiness in multiple summaries, this paper investigates whether predictors of utterance-level prominence extend to the discourse level, and how they interact across 24 spoken and written genres of English. We examine features including grammatical function, definiteness, entity type, linear order, discourse relations and hierarchy, and referential structure, as well as the impact of genre. Our results show that utterance-level predictors significantly correlate with discourse-level salience, but interact with and are modulated by entity-level factors such as frequency and dispersion across the document. Multifactorial models reveal that no single factor determines salience; rather, discourse-structural and semantic features prove more robust than morphosyntactic ones, with substantial variation by genre and communicative intent.}
  \communicated{...}
  \dedication{...}
  \received{...}
  \accepted{...}
  \journalname{...}
  \journalyear{...}
  \journalvolume{..}
  \journalissue{..}
  \startpage{1}
  \aop
  \DOI{...}

\maketitle
\renewcommand*{\thefootnote}{\arabic{footnote}}
\section{Introduction} 
\vspace{-6pt}

It is a truism that not all parts of a text are created equal: some parts are more central to the main message and others more tangential, some facts or participants mentioned are more memorable than others, and if we had to boil down a text to just a short summary, we would be more inclined to cut out some content while retaining what we feel is most important or salient. While this proposition is probably uncontroversial, operationalizing any of the parts needed for studying it in detail certainly is: what are the `parts' of a text, who are `the participants', what does it mean to be memorable or salient and how would we reduce a text to its most important content in a consistent way? And if we could resolve all these issues, what could we learn about what drives salience at the discourse level? The goal of this paper is to answer the last question by first tackling these definitional challenges, and then examining a large amount of data on salience in different types of text. To foreshadow some of the directions we will pursue, this study focuses on entity terms, i.e.~referring expressions, as possible `participants' in a text (including humans, inanimate objects and even abstract notions), uses hierarchically nested discourse units in the tradition of Rhetorical Structure Theory (RST, \citealt{MannThompson1988}) as minimal `parts', and adopts a Computational Linguistics methodology using graded summary-worthiness as a metric for discourse-level salience (cf. \citealt{boguraev-1997-salience}), which we correlate with a range of properties of mentions in order to test hypotheses and build predictive models.

Previous work on salience has tackled concepts which can be divided into two  ostensibly unrelated strands: salience at the utterance level, identifying the most prominent participants in a proposition (e.g.~in Centering Theory, \citealt{grosz-etal-1995-centering,poesio-etal-2004-centering}), and discourse-level salience, roughly corresponding to identification of discourse topics \citep{vandijk1977sentencetopic,Charolles2020} or global relevance (\citealt{vandijk1979relevance}, sometimes called `global idea' or `global topic', \citealt{Guijarro2005}). Although this paper will focus on the latter notion, which we will refer to as \textbf{discourse-level salience}, we note that many of the properties that are assumed to promote utterance-level salience may also be relevant to discourse-level salience, and the question whether the two are connected or correlated is an empirical one.

For example, previous work on utterance-level salience (see \citealt{fillmore1977topics} for early work and \citealt{HeusingerSchumacher2019,BoswijkColer2020} for recent overviews)
has noted that salient entities are often expressed as grammatical subjects \citep{TomlinMyachykov2019}, tend to refer to animate, and especially human participants \citep{prat-sala2000discourse}, are more often realized as pronouns \citep{kaiser2005salience} or other definite or information structurally given expressions (see the Mental Salience Framework in \citealt{Chiarcos+2011+105+140,chiarcos-2011-evaluating}), appear earlier in sentences \citep{whiteley2024order}, and may be less predictable or evoke higher levels of surprisal \citep{zarcone2016salience,LinZeldes2026-surprisal}. To illustrate this, consider the ordinary sounding \ref{ex:prototypical-salient}, and contrast it with the more unusually phrased \ref{ex:unprototypical-salient}.

\ex. The kids played with a ball in a park next to some trees until 5 p.m.\label{ex:prototypical-salient}

\ex. In the park, 5 p.m. was the time some kids played ball next to some trees.\label{ex:unprototypical-salient}

From a local perspective, the biases mentioned above would suggest that the kids are the most salient entity in the context of the utterance in \ref{ex:prototypical-salient} -- they are animate, definite, realized early and as the subject, etc., whereas \ref{ex:unprototypical-salient} fronts inanimate participants (the park, 5 p.m. as the subject) to raise their relative prominence based on the markers proposed in the literature above. Although it is impossible to be sure about the discourse-level salience of the same participants without context, we may nevertheless suspect that \ref{ex:prototypical-salient} is part of a story about some kids in which the park and time of day are tangential, while the more unusual sounding \ref{ex:unprototypical-salient} would fit better if the time and place i.e.~`the park' and `5 p.m.' are salient topics in the discourse beyond the local context, for example because they relate to an alibi question. At the same time, discourse and utterance-level salience are not the same thing, and while some properties may be indicative of both, we will see below that others do not even apply across the two settings. To name one such important property, the number of mentions of an entity across the discourse is likely to be highly correlated with its overall salience \citep{givon1983topic}, but it cannot be quantified at the utterance level for an entity's first mention, when the reader or hearer does not yet know how often it will be mentioned later on.

Methodologically, many of the postulated univariate correlations between salience (especially utterance-level) and animacy, subjecthood, definiteness, discourse structure and more, have relied primarily on either non-quantifiable introspective approaches, or on experimental evidence using artificial stimuli and definitions of salience that cannot be applied consistently to language `in the wild'. The smaller subset of corpus studies in this area have often suffered from less robust and sometimes circular definitions of salience based on text-internal criteria that correlate with the linguistic variables of interest, and have never explored language across a broad range of text types. In this paper we aim to close these gaps by providing a clearly defined operationalization of discourse-level salience that is explicitly \textbf{extra-textual} and applicable to \textbf{any kind of text type}, which is crucial since many factors explored in previous studies vary broadly with genres.

We take a corpus-based approach rooted in definitions of discourse-level salience as it is understood in Computational and Psycholinguistics literature, seeking to identify the most important, meaningful and memorable parts of given language data, in ways that generalize to a broad range of written and spoken text types in English. The main contributions of this paper are:

\vspace{6pt}
\begin{itemize}
    \item 
    We propose a graded, extra-textual operationalization of discourse-level salience grounded in summary-worthiness, which enables fine-grained modeling of entity prominence across a broad range of genres.
    \item We present the first comprehensive analysis of discourse-level entity salience using richly annotated corpus data of both written and spoken English, including referential structure, discourse structure, and morphosyntactic features.
    \item We test hypotheses about a range of predictors, including theoretical constructs predicting utterance-level salience such as Centering Theory.
    \item We develop and evaluate interpretable regression models and other predictive machine learning models that identify the most robust indicators of salience and generalize well to out-of-domain text types.
\end{itemize}

\vspace{-4pt}
To guide our investigation, we ask the following research questions:

\vspace{6pt}
\begin{enumerate}
    \item What linguistic and discourse-level properties shape the graded salience of discourse entities? 
    \item How do these properties interact across genres?
    \item To what extent can corpus-based models of salience generalize across diverse discourse types, and what do their errors reveal about the nature of the phenomenon in context when they fail?
\end{enumerate}

The properties we will study include frequency and dispersion of entity mentions, linear position within documents and sentences, parts of speech and grammatical number, syntactic function (e.g.~subjecthood), entity types (including animacy), genre, as well as variables derived from Centering and RST, such as utterance-level salience rankings, transitions between sentences, discourse relations, presence of discourse markers and position within a discourse parsing graph structure.
\vspace{-10pt}

\section{Previous work}\label{sec:previous}
\vspace{-2pt}

Since this paper focuses on discourse-level salience, we refer readers interested in work on utterance-level salience to \citet{BoswijkColer2020}, which offers a meta-review of salience definitions in Linguistics outside of computational approaches, and \citet{HeusingerSchumacher2019}, who give a detailed discussion of definitions of utterance-level prominence with a focus on dynamic updates as discourse unfolds. 
In Computational Linguistics, salience in general has been seen as the ``degree of prominence or attention an entity is assigned'' \citep[23]{DBLP:conf/inlg/ChiarcosS04}. An influential early approach to utterance-level salience which will be relevant below is Centering Theory \citep{grosz-etal-1995-centering}, a ``framework for theorizing about local coherence and salience'' \citep[309]{poesio-etal-2004-centering}.

Centering proposes that discourse coherence is created by utterance-salient co-referring expressions that form chains of reference, such that certain types of linguistic marking (e.g.~subjecthood) lead listeners to assign more prominence to certain entities. Entities in every sentence are ranked by salience, creating an ordered set of `forward-looking centers' or \textit{Cf}, with the expectation that in a coherent text the following sentence will most likely refer back to the highest ranking member of Cf, often with a pronoun, creating the following sentence's `backward-looking center' or \textit{Cb} -- the most anaphoric element of the sentence. The theory categorizes transition coherence between sentences based on Cf and Cb: in a `Continuation`, the current Cb is the same as the previous Cb and also the highest-ranked element in Cf; in a `Retaining', the current Cb matches the previous Cb but is no longer the top element in Cf. In \ref{ex:centering}, adapted from \citet[211]{grosz-etal-1995-centering}, a pair of consecutive sentences illutrates the `Continuation' transition.
\vspace{-2pt}

\ex. [Susan]$_{Cf_1}^{}$ gave [Betsy]$_{Cf_2}^{}$ [a pet hamster]$_{Cf_3}^{}$. [She]$_{Cf_1}^{Cb}$ asked whether [Betsy]$_{Cf_2}^{}$ liked [the gift]$_{Cf_3}^{}$.\label{ex:centering}

{\noindent}The Cf ranking of entities in each sentence depends on a number of factors, including pronominalization (Cb > other pronouns > non-pronouns), grammatical function (subj > obj > other), and givenness (given/aforementioned > accessible/inferrable > new, cf. \citealt{GundelEtAl1993}), with ties broken by linear order (early > late). `Susan' ranks highest because she is the subject in the first sentence and no pronouns are used, followed by Betsy, an (indirect) object, and finally the hamster. In the second sentence, the only pronoun used, `She', is selected as the Cb, which links back to the previous sentence to maintain coherence. It is also ranked highest in the second sentence Cf ranking, and then Betsy, now also a subject, again outranks `the gift'. 

We emphasize these rankings apply strictly within utterances: we cannot infer relative discourse salience by comparing the Cf ranks of mentions from different sentences. However there are good reasons to suspect markers of utterance and discourse-level salience may be correlated: \citet[13]{givon1983topic} postulates correlations between grammatical devices encoding local topic identifiability and `thematic' (discourse level) structure, including cross-utterance factors such as distance to previous mentions, position within paragraphs, and `topic persistence', corresponding to the number of mentions in multiple consecutive sentences. Similarly, \citet[182--185]{Chafe1994} connected Giv\'{o}n's findings on grammatical correlates of discourse centrality (or `thematic importance') with activation costs: more topical constituents are more activated and can be evoked using reduced linguistic means, such as unstressed pronouns, which are most often subjects (see also \citealt{Ariel1990}). The extent to which Centering and other utterance-level notions are relevant to discourse-level salience is thus an open question, which seems well worth exploring.

The use of linear order in sentences as a `tie-breaker' is paralleled by previous work that has found that earlier mentions gain salience by forming a foundation for the interpretation of subsequent mentions. In sentence processing this has been called the Advantage of First Mention \citep[273--275]{Gernsbacher1997}, which promotes faster reading times and lexical access in post-hoc queries (e.g.~elicited recall of entities in a text). The effect persists when first mentions are non-subjects \citep{carreiras1995advantage}, meaning it is distinct from, and additive to the postulated prominence of subjects, at least in SVO languages such as English, or other subject-before-object languages (which covers over 83.3\% of languages in a sample of 1,376 languages in the World Atlas of Linguistic Structures, \citealt{wals-81}). Although work on linear ordering by \citeauthor{Gernsbacher1997} and others has mainly addressed ranking of salient referents within sentences, it is also possible that early ordering may be correlated with discourse-level salience, a hypothesis we intend to test in this paper.

Most empirical work on discourse-level salience comes from computational approaches to the practical task of Salient Entity Extraction (SEE, see \citealt{gamon2013identifying,lin-zeldes-2024-gumsley,bhowmik-etal-2024-leveraging}), which aims to identify the most salient entities in an arbitrary text based on labeled data. While early work focused on evaluating automatic algorithms, e.g.~by constructing an entity grid of all noun phrases with recurring mentions and identifying the spans of utterances that they cover using string identities or rule-based anaphora resolution \citep{lappin-leass-1994-algorithm,boguraev-1997-salience,BarzilayLapata2008}, more recent work has focused on human driven SEE. In these approaches, semantic denotations corresponding to clusters of mentions are the carriers of salience, rather than specific mentions in individual utterances, and analyses rely
on data either directly annotated by, or derived from human judgments of prominence or centrality.

In the most direct approach, annotators explicitly identify entities deemed important or central to a discourse \citep{dojchinovski2016crowdsourced,trani2018sel,maddela-etal-2022-entsum}, most often using binary (salient/non-salient) or ordinal labels (more/less salient, or Likert scales). Similar efforts have been applied to rating salience for propositions, as well as questions about propositions or entire texts \citep{wu-etal-2024-questions,trienes-etal-2025-behavioral}. While intuitively appealing, explicit salience judgments present methodological challenges: annotators may diverge significantly due to prior knowledge and individual biases, varied conceptions of importance, or subjective assessments about discourse relevance and its exact locus (\citealt{finlayson2015propplearner}; \citealt[15]{trani2018sel}). A further issue is the scalability of direct elicitation methods, particularly when annotating extensive data in a broad range of genres. 

More scalable and consistent alternatives to direct annotation have also emerged, including reliance on user behaviors such as click-stream data (e.g.~if an entity is relevant to a document, it will be clicked on in web searches for the entity, see \citealt{gamon2013identifying}), or pre-existing data structures, such as the presence of hyperlinks and category listings in Wikipedia \citep{wu2020wn}.\footnote{Another approach using neural model behaviors is notable in computer vision, where Salience Maps of the most relevant regions of images can be extracted from the pixels most crucial for object classification (\citealt{Szczepankiewicz2023}), though it has not yet been applied to linguistic SEE (we thank an anonymous reviewer for commenting on this).} These approaches are attractive because they are inherently independent of linguistic form and therefore allow us to examine which linguistic properties (e.g.~subjecthood, definiteness, etc.) correlate with salience, without introducing any circularity that may arise if we use linguistic form to rank salience globally in the way that Centering does locally (for example if we posit that entities realized as subjects more often are more salient than ones realized as non-subjects across all utterances). However these annotation-free approaches are also highly specific to certain kinds of data (web searches, Wikipedia) and task-based settings, rather than targeting general notions of salience. Importantly for our use case, it is unclear how they could be used to approach spoken data, or even long-form written text (e.g.~fiction). 

In response to the challenges above, recent research has advocated for summary-based approaches, which we employ below. These approaches rely on extracting one or more summaries for documents, and define salience in terms of \textbf{summary-worthiness}. 
First promoted by \citet{dunietz2014new} and developed further by \citet{lin-zeldes-2024-gumsley}, 
the basic idea is that, because summaries are naturally constrained to be short, they will only mention the most salient entities. Put inversely, if an entity is truly salient in a document, then it should be difficult to summarize its content without mentioning it. This approach has several advantages: 
\vspace{-6pt}
\begin{enumerate}
    \item It is extra-textual: since being salient does not depend on textual forms in the original document, we can safely study correlations between form and salience
    \item It is applicable to any text type, unlike relying on hyperlinks, click-data, etc. which are exclusive to Web data
    \item Given a summary and a list of entities in a document, it is relatively easy to agree whether each entity appears in the summary
\end{enumerate}
\vspace{-6pt}

 At the same time, the approach also has several potential pitfalls: 

\vspace{6pt}

\begin{enumerate}
    \item Humans may still disagree on the presence of certain entities in summaries
    \item Even if they do agree, summaries themselves introduce subjectivity and extraction bias (what makes it into a summary and how it is portrayed)
    \item Although it is hard to summarize a document without its most salient entities, it is comparatively easier to include non- or less salient entities
    \item Unlike click-data, which is quantitative, being mentioned is binary, prohibiting a quantitative notion of relative salience
\end{enumerate}

\vspace{-3pt}

For the first pitfall, it is crucial to show that humans can agree about whether or not entities have been mentioned in a summary, which we refer to as \textit{entity-summary-alignment}, and evaluate in the next section. Pitfalls 2--4 can all be mitigated substantially by using not one, but several summaries \citep{lin-zeldes-2025-gum}. We assume that even if one summary might miss something important or mention something tangential by chance or due to bias, this will not recur systematically in other summaries.\footnote{We thank an anonymous reviewer for noting that if the text itself is controversial, bias may lead to more subjectively skewed summaries. We recognize this limitation for highly charged texts, but do not foresee a substantial risk for our data, presented in Section \ref{sec:data}.} Below, we therefore use five summaries per document and operationalize salience as \textbf{the number of summaries in which an entity is mentioned}, ranging from 0 to 5. This means that subjectivity in one summary is balanced by others (adressing pitfall 2), less salient entities are unlikely to appear many summaries (3), and the resulting data is graded, providing relative scores (4).

\vspace{-12pt}
\section{Data}\label{sec:data}
\vspace{-8pt}


To get a comprehensive view of what is salient in different kinds of texts while also having access to annotations relevant to our research questions, we use the Georgetown University Multilayer corpus (GUM, \citealt{Zeldes2017}), a freely available English dataset annotated for dependency syntax using Universal Dependencies \citep{de-marneffe-etal-2021-universal}, nested named and non-named entities with coreference resolution in the Universal Anaphora format \citep{poesio-etal-2024-universal}, discourse parsing using Enhanced Rhetorical Structure Theory (eRST, \citealt{ZeldesEtAl2025eRST}), and more. 

The corpus covers a total of around 268,000 tokens in over 280 documents, divided into four partitions: training, development and test partitions for the main corpus, and a second out-of-domain test partition called GENTLE (GENre Tests for Linguistic Evaluation, see \citealt{aoyama-etal-2023-gentle}) with especially challenging genres. Table \ref{tab:gum-gentle} gives an overview of document types, sources and amounts of data.

\begin{table}[tbh]
\centering
\resizebox{\textwidth}{!}{%
\begin{tabular}{llrrllrr}
\toprule
\textbf{Genre} & \textbf{Source} & \textbf{Docs} & \textbf{Tokens} & \textbf{Genre} & \textbf{Source} & \textbf{Docs} & \textbf{Tokens} \\
\midrule
Academic writing      & Various            & 18 & 17,169 & How-to guides       & wikiHow      & 19 & 17,081 \\
Biographies           & Wikipedia          & 20 & 18,213 & Interviews          & Wikinews     & 19 & 18,196 \\
Vlogs                 & YouTube   & 15 & 16,864 & Letters             & Various      & 12 & 9,989  \\
Conversations         & UCSB Corpus        & 15 & 17,932 & News stories        & Wikinews     & 24 & 17,186 \\
Courtroom transcripts & Various            & 9  & 11,148 & Podcasts            & Various      & 10 & 11,986 \\
Essays                & Various            & 9  & 10,842 & Political speeches  & Various      & 15 & 16,720 \\
Fiction               & CC fiction         & 19 & 17,511 & Textbooks           & OpenStax     & 15 & 16,693 \\
Forum discussions   & reddit             & 18 & 16,364 & Travel guides       & Wikivoyage   & 18 & 16,515 \\
\midrule
\textbf{Total GUM:}    &                    & \textbf{255} & \textbf{250,409} & \multicolumn{4}{l}{(test: 32 docs; dev: 32 docs; train: 191 docs)} \\
\midrule
Dictionary entries    & Wiktionary         & 3  & 2,423  & Poetry              & Wikisource   & 5  & 2,090  \\
Esports commentaries  & YouTube            & 2  & 2,149  & Mathematical proofs & ProofWiki    & 3  & 2,106  \\
Legal documents       & Wikisource         & 2  & 2,288  & Syllabuses          & GitHub       & 2  & 2,431  \\
Medical notes         & MTSamples          & 4  & 2,164  & Threat letters      & casetext     & 5  & 2,146  \\

\midrule
\textbf{Total GENTLE:} &                    & \textbf{26}  & \textbf{17,799}  & 
\multicolumn{4}{l}{(test only: 26 docs)} \\
\midrule
\textbf{Grand total:}&                    & \textbf{281} & \textbf{268,208} & \multicolumn{4}{l}{} \\
\bottomrule
\end{tabular}%
}
\caption{Overview of data in GUM (top) and GENTLE (bottom).}
\vspace{-20pt}
\label{tab:gum-gentle}
\end{table}

Single human-written summaries were initially collected for the corpus by \citet{liu-zeldes-2023-gumsum}, which were subsequently expanded to five manual summaries per document for the dev/test sets, and four LLM summaries for the remaining data next to a single human-written one (see \citealt{lin-zeldes-2025-gum}).
All summaries were then manually corrected and aligned using the GitDOX annotation interface \citep{ZhangZeldes2017} to indicate which entities in each document are mentioned in which summary, allowing for the extraction of salience scores as described above. To give an idea of human agreement levels on entity-summary-alignment, we carried out a double annotation experiment 
over a test set of 3,283 entities and 24 summaries. Using gold-standard entity annotations as candidates, annotators reached a very high agreement of $\kappa=0.9782$ on binary judgments of whether each entity appeared in the summary, showing that, in our setting, pitfall 1 is mitigated: humans can reliably agree on the presence of entities in summaries.

The sixteen genres in the GUM partitions cover both spoken and written text types, including cross-modal types such as speeches (often written to be spoken), and more or less spontaneous modes of speech, from spontaneous face-to-face conversations to often more scripted YouTube vlogs. Written data ranges from formal academic writing, to user generated content from Reddit, as well as instructional genres such as how-to guides from Wikihow and travel guides from Wikivoyage. The additional GENTLE partition adds eight diverging and potentially challenging text types: dictionary entries which are metalinguistic by nature, live eSports commentary which is highly situated and unscripted depending on spontaneous video game action, as well as more formal medical notes, legal documents, course syllabi, poetry, mathematical proofs and finally, threat letters. For most of this paper, we will use the train and dev partitions to study salience, before moving on to  testing multifactorial models on the test partitions in Section \ref{sec:multifactorial}. Since data in the normal test partition comes from the same sources as in train, we may expect models to perform particularly well on it; we therefore also evaluate models using the GENTLE partition, with the idea that generalization to those text types may offer strong support for the relevance of the associated models.
\vspace{-12pt}

\section{Methods}

With our salience-annotated data in hand, we aim to answer a number of research questions about salience in naturally occurring data. In the following section, we will focus on individual linguistic sub-systems, such as morphosyntactic structure (especially subjecthood, but also part of speech, singular versus plural number, definiteness and linear order within sentences), animacy and entity types (Section \ref{sec:people}), discourse structure in the framework of eRST (including discourse markers, discourse functions, and the hierarchy of discourse units, Section \ref{sec:rst}) and referential structure in the framework of Centering Theory (Section \ref{sec:centering}). In each of these analyses we will provide descriptive statistics on categories correlating with discourse-level salience, and construct linear models to compare the significance of individual categories, testing hypotheses about the relevance of different factors. In Section \ref{sec:multilayer}, we will turn to multilayer analyses which incorporate significant features from the more targeted models to build the most accurate and generalizable predictive model possible, which we will test on both unseen in-domain data and out-of-domain data from the GENTLE subset described above. Here we will follow the objectives of testing whether or not predictors from single layer analyses remain significant when all features are combined (Section \ref{sec:multifactorial}), and extract profiles of more and less discourse-salient entities, with a focus on genre variation (Sections \ref{sec:genre}--\ref{sec:clustering}). One of our objectives throughout will be to test whether features connected with utterance-level salience in previous work are also predictive of discourse-level salience, and how robust these are in different types of text.

In all models below, we will distinguish between \textbf{instance-based} variables, which apply to individual entity mentions in a specific sentence, versus \textbf{entity-based} variables, which aggregate data about entities from all of their mentions, and for which we therefore have one value per unique entity (i.e.~a cluster of mentions). For example, the grammatical function of a mention (subject, direct object, etc.) is instance-based, since it can change in subsequent mentions of the same entity. By contrast, the number of mentions an entity has across an entire document is a single scalar value associated with the entire cluster of all of its mentions. 

To construct our regression models, we use implementations of mixed effects models in R. Since salience labels are discrete (values can only be integers, 0--5, with no intermediate values) and follow a U-shaped, rather than a normal distribution (salience=0 applies to over 62\% of the mentions, and nearly 17\% of mentions have salience=5, see below), we employ beta-binomial regressions with library \texttt{glmmTMB}, unless otherwise stated. This type of regression is appropriate for aggregate U-shaped integer data: we model our salience scores as the sum of five binomial success `chances', corresponding to a positive salience vote based on each summary. In Section \ref{sec:multifactorial} we will also employ tree ensemble models using the \texttt{ranger} libary, specifically applying the Extra Trees algorithm (\citealt{GeurtsErnstWehenkel2006}, see Section \ref{sec:multifactorial} for more details). Finally, to complement our formal hypothesis testing with qualitative and graphical analyses, we will extract profiles of salient and non-salient entity types. We use  t-SNE (t-distributed Stochastic Neighbor Embedding, \citealt{MaatenHinton2008-tsne}), which is a bottom-up technique prioritizing local similarities to produce a low dimensional plot of data points based on their proximity in high dimensional space. We feed the variables from all of our analyses to the model to produce a model space, which we analyze with qualitative examples from different regions of the space in Section \ref{sec:clustering}. A glossary with relevant terms, variable names and label values is provided in the Appendix.

\vspace{-12pt}

\section{Single layer analyses}

\vspace{-6pt}
\subsection{Is discourse salience marked through subjecthood?}\label{sec:subj}
\vspace{-3pt}

In previous work, the most frequently invoked characteristic of utterance-level salience has been a bias towards realization as the grammatical subject (including via passivization if needed). However, a number of other morphosyntactic factors that are often collinear with subjecthood and each other have also been studied frequently, such as pronominalization, short mention length, and definiteness or givenness. Partly orthogonal to the grammatical categories of word forms is linear order, either within the sentence or the entire document, with early order in the sentence again being correlated with subjecthood in English. Using our data, we can first test the extent to which the two most obvious confounds of subjecthood, pronominalization and linear order, correspond to salience. For pronominalization we can run an instance-based t-test, which, given the size of the data ($N\sim53,000$ in the training partition) is unsurprisingly significant, but with only a medium effect size ($p<0.00001$, $g=0.65$). For linear order, we observe that position of the first mention of an entity (expressed as the ratio of the entity-based first token index/length of the document) is much better correlated with salience score than linear positions within a sentence (instance-based position in sentence/sentence length) -- document position exhibits Pearson's $r=-0.251$, while position in the sentence has only $r=-0.059$ (both results highly significant, $p<0.00001$). While these results are all clearly significant, we would now like to know: a. whether grammatical function, and in particular subjecthood, is also significantly associated with discourse-level salience, and b. whether this effect is significant even when position, pronominalization and other form-based factors are known.

\begin{figure}[h!tb]
\centering
\includegraphics[width=1\textwidth]{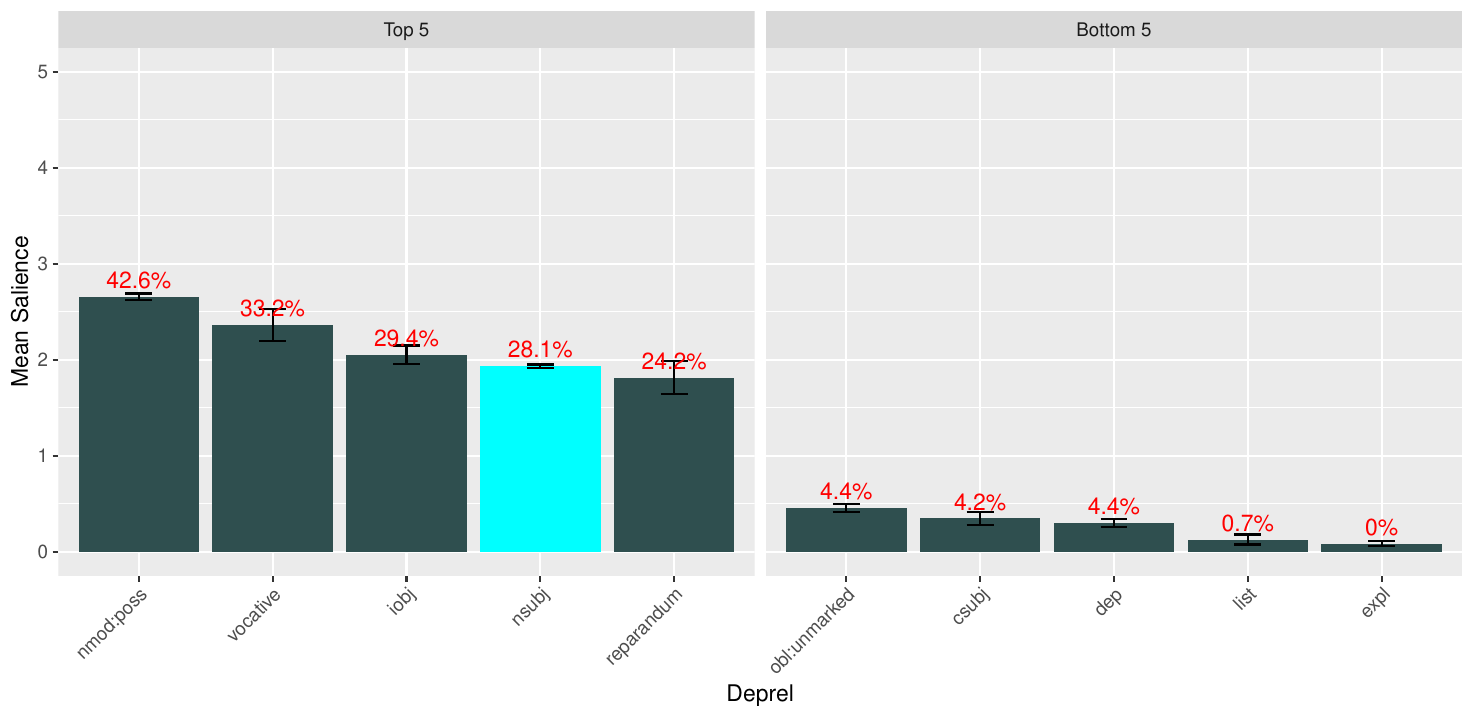}
\caption{Mean salience scores for mentions with the top and bottom five dependency relations. Error bars reflect GLHT-adjusted 95\% confidence intervals and the percentage of mentions with $salience=5$ is listed in red; the bar for nominal subjects is highlighted.}
\label{fig:deprels}
\vspace{-9pt}
\end{figure}

To answer the first question, we first examine grammatical functions in isolation. Since function labels are an instance-based property, we investigate this again at the mention level using the training partition, meaning that entities with more mentions will have an outsize effect on results; at the same time, we note that \textbf{singletons}, i.e.~entities mentioned only once in a text, are the largest part of our data, covering over 72\% of entities, and 37.3\% of mentions. Figure \ref{fig:deprels} gives the mean salience score for the top and bottom 5 most/least salient dependency relations using the UD tagset, overlaid with their percentage of entities scoring a salience rating of 5, as well as adjusted 95\% confidence intervals.\footnote{Since we compare 45 attested UD labels, confidence intervals are adjusted using General Linear Hypothesis Testing (GLHT, \citealt{HothornEtAl2008}) to compensate for compounded alpha errors. Any pair with non-overlapping error bars differs significantly for $\alpha=5\%$. 
}

As the figure shows, nominal subjects (\texttt{nsubj}) are indeed much more salient than almost all of the other grammatical functions in the corpus. However, they actually rank fourth out of the total 45 labels used in the data for any entity, behind \texttt{nmod:poss} (possessive pronouns and genitive 's modifiers), \texttt{vocative} and \texttt{iobj} (indirect objects, i.e.~ditransitive recipients); in the case of possessives and vocatives, the difference is also highly significant, and the proportion of score=5 salient mentions is 52\% and 18\% higher for possessives and vocatives than it is for subjects -- a very substantial difference. 

Moving to the second question above, although we have already seen that linear order and pronominalization are significant correlates of salience, we find that the effect of grammatical function easily remains significant even in a linear model incorporating all of these features. 
As additional form-based predictors, we include not only a binary pronominalization feature, but instead consider the categorical part-of-speech of each mention's head word (with \textsc{PRON} being one of 16 attested values\footnote{The 17\textsuperscript{th} possible tag \texttt{PUNCT} is never used as the head of an entity.} using Universal POS tags, \citealt{petrov-etal-2012-universal}), as well as definiteness and singular number as binary variables, both of which have been positively linked to utterance-level salience in previous work. As the instance-based model shown in Table \ref{tab:anova-salience} demonstrates, all form-based variables remain significant.\footnote{Since there are 45 dependency relation labels, we provide significance values based on Likelihood-Ratio tests for single term deletions of the complete model, and follow the same practice in models below.}

\begin{table}[ht]
\centering
\caption{ANOVA table for model: \texttt{salience \textasciitilde ~ position\_in\_sent + position\_in\_doc + dependency\_relation + part-of-speech + definiteness + number}, with predictors sorted by change in AIC based on single term deletions, from the least to the most informative.}
\label{tab:anova-salience}
\begin{tabular}{lrrrrr}
\toprule
Term & Df & Deviance & AIC & F value & Pr($>$F) \\
\midrule
\textless{}none\textgreater{}    &     & 167727 & 212799 &         &                      \\
position\_in\_sent &  1 & 167883 & 212847 &   49.748   & $1.769 \times 10^{-12}$ *** \\
definiteness           &  1 & 168095 & 212914 &  117.161   & $< 2.2 \times 10^{-16}$ *** \\
position\_in\_doc              &  1 & 169808 & 213456 &  661.929   & $< 2.2 \times 10^{-16}$ *** \\
number             &  1 & 171418 & 213960 & 1173.888   & $< 2.2 \times 10^{-16}$ *** \\
part-of-speech          & 15 & 174003 & 214731 &  133.062   & $< 2.2 \times 10^{-16}$ *** \\
dependency\_relation       & 44 & 174642 & 214869 &   49.979   & $< 2.2 \times 10^{-16}$ *** \\
\bottomrule
\end{tabular}
\vspace{-9pt}
\end{table}

In other words, while the answer to the question `is discourse salience marked via subjecthood' cannot be a simple `yes' -- other functions are in fact stronger correlates -- grammatical functions matter, and subjects are among the more salient. Looking at the data qualitatively, one property that the most salient grammatical functions have in common quickly becomes clear -- as shown in examples \ref{ex:voc}--\ref{ex:poss}, they tend to represent human participants: (relevant entities are given in brackets with the salience score in a subscript and the dependency relation in superscript)

\ex. \textit{``It doesn't hurt,} [\textit{Gram}]$_{5}^{vocative}$\textit{,''} [\textit{Renata}]$_{5}^{nsubj}$ \textit{says}\label{ex:voc} 

\ex. [\textit{the administration}]$_{0}^{nsubj}$ \textit{asked} [\textit{me}]$_{5}^{iobj}$ \textit{for} [\textit{a job description}]$_{0}^{obl}$\label{ex:iobj}

\ex. [\textit{Johann}]$_{2}^{nsubj}$ \textit{also plagiarized ... from} [[\textit{Daniel's}]$_{5}^{nmod:poss}$ \textit{book}]$_{1}^{obl}$\label{ex:poss}

In \ref{ex:voc}, a character in a short story is addressing another, and both are mentioned in all summaries. Although subjects are prominent and main characters are often subjects, not all subjects are characters, while vocatives almost guarantee a human participant. In \ref{ex:iobj}, taken from a political speech, a minister is telling what the administration asked of her -- the minister is salient and mentioned in summaries of her speech, but the administration is not mentioned in the summaries, leading to a score of 0. Finally in \ref{ex:poss}, taken from a biography of Daniel Bernoulli, we see that Johann Bernoulli, his father, is a subject, and somewhat salient (mentioned in 2 summaries), but not as much as Daniel. 



But if the most salient entities in documents are actually vocatives, indirect objects and possessives, rather than subjects, why have these functions enjoyed so much less attention? To better understand this, consider how many of the mentions in the three grammatical categories outranking subjects for salience have the \textsc{person} entity type. In our training data, indirect objects consist of mentions of people 82\% of the time, vocatives 94\%, and possessors are at about 80\%. By contrast, subjects only represent people 59\% of the time. Seen from this perspective, what is unique about subjects is not that they are the most discourse-salient grammatical function, but rather how salient they are despite having a lower proportion of \textsc{person} entities. In other words, while most possessors and almost all vocatives are people as in \ref{ex:poss} or \ref{ex:voc}, far from all subjects are. 

Given the overwhelming prevalence of human participants as salient entities in the most salient grammatical functions, and with the qualitative examples above in mind, the next question we address is whether mentions of people, or perhaps more broadly animate entities, are the best (and simplest) explanation for what drives discourse-level salience scores?
\vspace{-9pt}

\subsection{Are people simply the most salient?}\label{sec:people}
\vspace{-6pt}
It has been noted that as humans, we tend to focus on mentions denoting people at the expense of other participants. For example, animate referents, both people and to an extent animals, are recognized more quickly in lexical decision tasks \citep{Bonin_Gelin_Dioux_Meot_2019}, are recalled better in episodic memory \citep{gelin2017animacy}, and are classified more quickly and accurately in visual tasks \citep{new2007category}. Animacy effects are also known to be strong enough to override other constraints, such as the agent-first preference, with human participants often being promoted to sentence-initial position as subjects, even for predicates whose word-order preferences are violated as a result \citep{Ferreira1994ChoiceOP}. Animacy is sometimes treated as a binary distinction (humans/non-humans), but is also often viewed as a scale, where people outrank animals \citep{deSwartdeHoop2018,sorlin2018anthropocentrism}, and some authors distinguish further levels: 1st/2nd person outranking 3rd person, proper names outranking common noun entities, inanimates at the bottom \citep{croft2003typology}, sometimes additionally with concrete objects above abstracts \citep{siewierska2004person}.

\begin{figure}[tbh]
\centering
\includegraphics[width=0.9\textwidth]{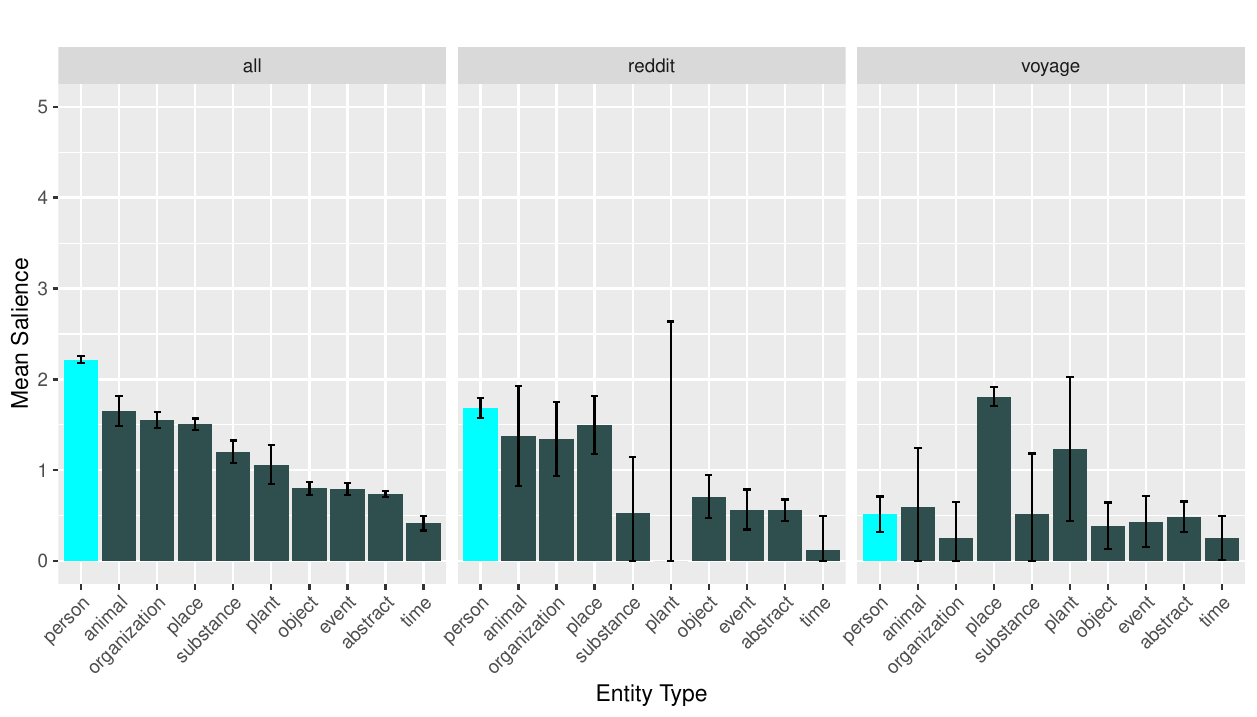}
\caption{Mean salience scores for mentions in each entity type for the entire corpus (left), reddit forum discussions (middle) and Wikivoyage travel guides (right). Error bars reflect GLHT adjusted 95\% confidence intervals and the bars for \textsc{person} entities are highlighted.}
\label{fig:etypes}
\vspace{-9pt}
\end{figure}

With this in mind and given the examples in the previous section, it seems reasonable to suspect that \textsc{person} (and perhaps \textsc{animal}) entities should be more salient on average than other types -- we can refer to this proposal as the Animate Salience Hypothesis (ASH). Figure \ref{fig:etypes} gives the mean salience score per mention by entity type, using the 10 types distinguished in the GUM corpus. The first panel on the left shows the overall distribution, and clearly confirms the ASH proposal. It is also in line with a graded runner-up position for \textsc{animal} entities, while placing the most abstract categories, \textsc{abstract} and \textsc{time}, at the bottom. 
Surprising rankings compared to animacy hierarchies include places outranking inanimate objects, and plants not being third, despite being living things. 

\vspace{-12pt}


Although the overall picture that emerges from the first panel supports an ASH account, the question remains whether this is always so, or are there kinds of texts in which different biases apply? To answer this question, we rank the genres in the corpus by the Spearman correlation of their entity distributions with the overall means from the first panel. As it turns out, and somewhat surprisingly, no single genre in the corpus follows the overall ranking exactly. We find that reddit forum discussions (middle panel)  follow the overall ranking most closely, with \textsc{person} in the lead (highlighted in cyan), \textsc{animal, organization} and \textsc{place} next (all not significantly different in either panel) and only \textsc{substance} and \textsc{plant} deviating strongly, in large part because plants are almost never mentioned in this subcorpus and are never salient, leading to a large error bar and a mean of zero. 

The most deviating genre (lowest Spearman correlation with the overall ranking), by contrast, is travel guides from Wikivoyage (right panel), in which \textsc{person} entities rank only fourth in salience, behind \textsc{place}, \textsc{plant} and \textsc{animal}. We emphasize that this data does not hinge on \textbf{frequency} but on \textbf{likelihood of salience given a mention}, meaning it should not matter in the plot whether people are mentioned frequently in travel guides or not -- only whether they are likely to be salient once mentioned. In sum, Figure \ref{fig:etypes} shows that while people 
are overall more likely to be salient, the picture can change substantially across genres. It may not be surprising that places are salient in travel guides, but if we want to model what makes entities salient across a broad range of text types, we must take this, too, into account, and we cannot simply assert that `people are more salient'.


So why are people not so salient in travel guides, even when mentioned? In many cases, the answer is that by the very nature or travel guides as a genre, people are likely to be mentioned only in passing, and they are often ultimately not `summary-worthy'. To see why, consider examples \ref{ex:place-not-person1}--\ref{ex:place-not-person2} from two travel guides.

\ex. [\textit{Sydfynske Øhav}]$_{5}^{place}$ \textit{is located south of} [\textit{the Danish island of Funen}]$_{4}^{place}$ \textit{... with} [\textit{a bit more than 20,000 inhabitants}]$_{1}^{person}$ \textit{in total}.\label{ex:place-not-person1}

\ex. [\textit{Oakland}]$_{5}^{place}$ \textit{nurtured novelists} [\textit{Amy Tan}]$_{0}^{person}$ \textit{and} [\textit{Maya Angelou}]$_{0}^{person}$\label{ex:place-not-person2}

In \ref{ex:place-not-person1}, the population size of a place is not as salient as the place itself and is picked up by only a single summary ($salience=1$), indicating the number of people at a destination is more tangential than, for example, another more prominent landmark (the position relative to another island). In \ref{ex:place-not-person2}, Oakland itself as the subject of the guide is easily the most salient entity, while background information about authors who lived there is unlikely to be the focus of a longer text about the city. This suggests an important distinction between \textbf{main points} and \textbf{supporting information}, which depends on the communicative goals of the text.

But if this is a key distinction, it begs the question: can we distinguish the tangential or background parts of a text  from the main points or objectives in the structure of the document in a theoretically well-founded, operationalizable way? If so, would our operationalization consistently predict discourse-level entity salience, and how would it interact with other factors? We turn to these questions next.

\subsection{Does discourse structure signal salience?}\label{sec:rst}

\begin{figure}[h!]
\centering
\includegraphics[width=1\textwidth, trim={5.8cm 3.3cm 5.37cm 11.6cm}, clip]{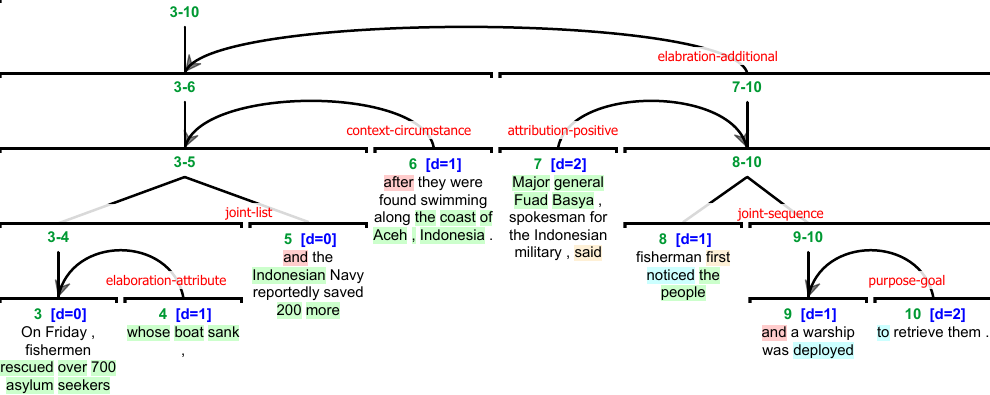}
\caption{An eRST graph fragment. The coordinate EDUs 3 and 5 are the most central (depth \textit{d=0}) since all other units point to them directly or indirectly. Signal tokens are highlighted, e.g.~explicit DMs like `and' or `after' are marked in red.}
\label{fig:erst}
\vspace{-9pt}
\end{figure}

So far we have focused on the properties of referring expressions themselves, but it seems clear that salience is also signaled by properties of the usually verbal predications they participate in -- how central they are to the main point of the discourse, and what function they serve with respect to other propositions (e.g.~tangential elaborations may be less salient environments than predicates providing proximal causes for events). To categorize the discourse functions of predications and quantify their centrality, we can use the Enhanced Rhetorical Structure Theory (eRST) graphs available in GUM/GENTLE, an extension of Rhetorical Structure Theory (RST, \citealt{MannThompson1988}) which is described in \citet{ZeldesEtAl2025eRST}. The eRST graphs divide each document into Elementary Discourse Units (EDUs, roughly corresponding to predications or clauses), and create a hierarchical structure in which each EDU receives a function label, or else heads a block of EDUs with a corresponding label. Figure \ref{fig:erst} illustrates a fragment from such a graph using the rstWeb tool \citep{gessler-etal-2019-discourse}.

In the figure, the two most central EDUs are the coordinate units [3] and [5], which form a \textsc{joint-list} relation. These units provide the main information in the passage: 700 hundred asylum seekers were rescued by fishermen (unit [3]) and 200 more by the Indonesian navy ([5]). The centrality of these units can be noticed by the fact that all other units point to them (cf.~\citealt[115]{Stede2012}), directly (as [4] points to [3] directly as a type of \textsc{elaboration}, specifying that their boat sank) or indirectly (the comment by Fuad Basya in units [7]--[10] is a type of \textsc{elaboration} on the whole block of [3]--[6]). We can treat the number of relation `hops' (directed arrows) away from the most central unit as a measure of discourse nesting depth -- EDUs [3] and [5] have depth 0 ([d=0] in the figure) since they are the most central, unit [4] has a depth of d=1, and so on. 

Figure \ref{fig:erst} also provides the relation labels, which have two hierarchical levels (units [3] and [5] are a temporally unordered \textsc{joint-list}, while [8] and [9] are connected by a temporally ordered \textsc{joint-sequence}), as well as highlights for tokens that are annotated as signals of a particular relation. For example, the \textsc{joint-list} relation connecting [3] and [5] is signaled by the word `and', an explicit discourse marker (DM), highlighted in red in unit [5]. Other types of signals include lexical signals, such as `said' (unit [7], in yellow), which signals an \textsc{attribution}, semantic signals (in green) such as repeated mention indicating an \textsc{elaboration} (e.g.~`the people' in [8] refers back to the asylum seekers, signaling the \textsc{elaboration-additional} relation) and syntactic signals highlighted in cyan, such as the `to'-infinitive signaling the \textsc{purpose} relation in unit [10]. The source of a directed relation is referred to as a \textbf{satellite}, and the unit it points to is the \textbf{nucleus} (e.g.~unit [10] is a \textsc{purpose} satellite whose nucleus is unit [9]).

Using these annotations, we can investigate several potential correlates of salience. First, we would like to know whether centrality, operationalized as depth in the tree, is correlated with salience. We note that, although we may expect central units in the eRST graph to overlap with content mentioned in summaries (and in fact, RST tree positions have been used as inputs to extractive summarization algorithms, see \citealt{marcu1997rhetorical,Stede2008RSTRevisited,UzedaEtAl2010}), the relationship between tree position and our salience metric is non-trivial: trees were generated without access to the summaries, and our object of study is not the propositions represented by the units in the tree, but rather their participating entities. The same salient entities can easily appear both in high and low-depth units, and completely non-salient entities often appear in central discourse units (for example tangential time and place adjuncts). Additionally, our summaries are abstractive and not extractive: many documents do not have an EDU or subset of EDUs that correspond well to the summary on a literal level, meaning the structure of the RST graph and the summary can differ considerably. The interactions between discourse tree depth, grammatical form and entity types may therefore be quite complex.

We hypothesize that entities whose head token is in a central unit like [3] (e.g.~`700 asylum seekers', headed by `seekers', with depth=0) will be more salient than entities in more nested units, such as `Fuad Basya' in [7] (depth=2), notwithstanding interaction with the features we have already studied (e.g.~asylum seekers and general Basya are both animate, but the latter is a subject while the former is a direct object). Since the maximum depth in a document can depend on the amount of EDUs, and therefore the length of the document, we normalize depths to percentiles (depth 0 remains 0, and 1 means the maximum depth in a document). There are then still two different ways of measuring the impact of depth: either at the mention level (each mention has its own depth) or at the entity level, for which we will use the depth of the first mention of each entity, i.e.~the centrality of the EDU in which an entity was introduced.

\begin{figure}[tbh]
\centering
\includegraphics[width=0.9\textwidth]{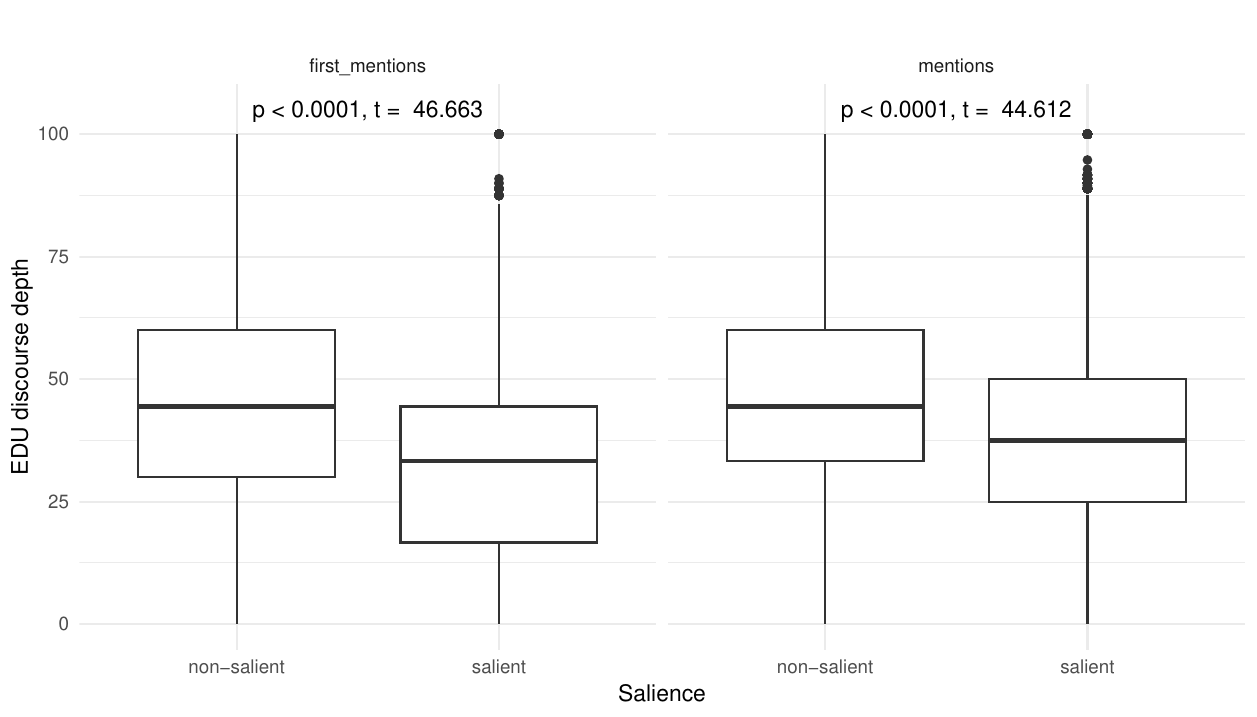}
\vspace{-10pt}
\caption{Box plots of EDU discourse depth for non-salient and salient entity first mentions (left) and all mentions (right). Entities in less deep EDUs are significantly more salient.}
\label{fig:edu-depth-boxplots}
\end{figure}

Figure \ref{fig:edu-depth-boxplots} provides boxplots for depth in non-salient (score=0) and salient (score>0) entities. While differences are significant in both cases, the effect is stronger for first mentions. Pearson correlations between salience and depth scores are also significant, and stronger for first mentions: $r=-0.151$ for all mentions versus $r=-0.235$ for first mentions (both $p<0.00001$). These findings strongly support the idea that regions of the text analyzed as more central in the eRST graph provide a boost to salience scores (or conversely, that the presence of salient entities corresponds to analysts assigning EDUs to a more prominent position in the graph when creating the eRST annotations).


\begin{figure}[bth]
\centering
\includegraphics[width=1\textwidth]{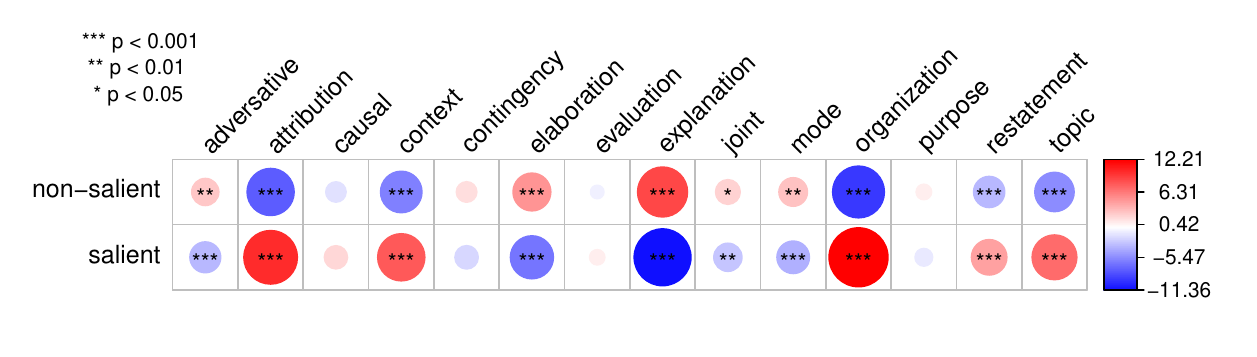}
\vspace{-26pt}
\caption{$\chi^2$ residuals for coarse discourse relation labels of discourse units containing salient and non-salient mentions. Colors show positive (red) and negative (blue) deviations from expected values and sizes represent absolute values. Asterisks show significance.}
\label{fig:rsd-label-residuals}
\vspace{-6pt}
\end{figure}

Turning to the discourse relation labels themselves, we would like to know whether there are also imbalances between functions such as \textsc{elaboration}, \textsc{purpose} etc. In the interest of space we use only the 15 top-level relation categories, and collapse subtypes such as \textsc{joint-list} and \textsc{joint-sequence} into \textsc{joint}. Figure \ref{fig:rsd-label-residuals} plots the chi-squared residuals of a contingency table comparing salient and non-salient entities across relation categories.


The Figure shows that some relations are neutral with respect to salience and do not have significantly lower or higher proportions of salience than expected, for example \textsc{evaluation}. Other relations, such as \textsc{elaboration}, \textsc{explanation} or \textsc{adversative} (for example concessions with `although') have significantly fewer salient entities than others. This is perhaps unsurprising, since elaborations, as in \ref{ex:elab-example}, tend to provide additional, often optional expansions, explanations as in \ref{ex:expla-example} by nature provide supporting information, and adversatives such as concessions tend to present information that is contrary to the main point being made (satellites are given in square brackets with the relation type in a subscript, and nuclei are in angle brackets). 

\ex. \textlangle\textit{I saw her sometimes:}\textrangle [\textit{In a bright pink dress...}]$_{elaboration}$ (fiction)\label{ex:elab-example} 

\ex. \hspace*{-0.15cm}\textlangle\textit{York is known as England's "City of Festivals"}\textrangle [\textit{as there are regular \\\hspace*{-0.15cm}cultural festivals every year.}]$_{explanation}$ (travel guide to York)\label{ex:expla-example}

By contrast, relations such as \textsc{attribution},  \textsc{organization} (used among other things to connect headings in text), \textsc{context} and \textsc{topic} (the category containing question-answer pairs), have significantly more salient entities than expected. This too can be explained intuitively: attributions present sources of information such as speakers of direct speech (unit [7] in Figure \ref{fig:erst}), which tend to be human participants -- this fits with the ASH findings in Section \ref{sec:people}. The centrality of entities appearing in headings is also unsurprising, as is the importance of entities about which questions are being posed -- questions are often regarded as a way of managing common ground \citep{klein1987quaestio,klein2002quaestio,roberts2012information,doering2018modal}, leading the topic of a conversation. The \textsc{context} class, which includes background information, is a little more surprising, since backgrounded information is typically less central; but background information blocks tend to be whole sentences, which often mention the salient entities about which background is provided, as in \ref{ex:background-example}, taken from a conversation. 

\ex. \textlangle\textit{And I saw this guy Nierman.}\textrangle [\textit{I never heard of him, at that time.}]$_{context}$\label{ex:background-example} 

Here the exact same entities are expressed in the nucleus sentence (the speaker saw Nierman) and the sentence providing the background context (he had not heard about Nearman before). This contrasts for example with \textsc{explanation} clauses, which are often subordinate clauses introduced by DMs such as `since' or `as', in which salient entities from a main clause may not be repeated as often, as in \ref{ex:expla-example}, where the salient city of York is not mentioned again in the \textsc{explanation} clause.

As for signals, we limit ourselves here to considering the effect of explicit relation marking using DMs: are entities introduced in EDUs signaled by explicit DMs more or less salient? On the one hand, explicit marking could correspond to more prominent EDUs, but on the other hand, if entities are salient, it is possible that the implications of the EDUs they are in are more obvious, requiring less explicit marking. Empirically we find that explicit relation EDUs significantly disfavor salient entities ($p<0.0001$, $\chi^2=23.292$) -- this is in line with previous findings about interactions between the likelihood of pronominal realization of entities and DM use, as well as the likelihood of DMs in cognitively complex environments \citep{patterson-kehler-2013-predicting,HoekZuffereyEversVermeulSanders2017}.


Although this finding appears to support the idea that increased salience allows for more implicitness in relation marking, it is easily possible that colinearities account for it instead -- implicit relations could tend to be less deeply nested (and they are, with a mean depth of 0.427 vs. 0.501, $t=39.927, p<0.00001$) or they could be associated with some of the relations that favor salience (this is only partly true: for example for the \textsc{organization} relation characteristic of headings, which corresponds to no DMs, whereas \textsc{context} relations are often marked by `when', `while', `before' or `after'). The same could be true of collinearities with EDU depth, and in particular, we might suspect that document position may be confounded with the degree of nesting, with early EDUs tending to be less nested. We should therefore consider whether the effects in this section all remain significant when combined in a single model, to which we also add the entity position in the document as a percentile start point to control for linear position. As Table \ref{tab:anova-salience-rsd} shows, all of the effects remain highly significant.

\begin{table}[ht]
\centering
\caption{ANOVA table for model: \texttt{salience \textasciitilde ~discourse\_relation + edu\_depth + position\_in\_doc + explicit\_DM}, with predictors sorted by change in AIC based on single term deletions, from the least to the most informative.}
\label{tab:anova-salience-rsd}
\begin{tabular}{lrrrrr}
\toprule
Term & Df & AIC & LRT & Pr($>$Chi) \\
\midrule
\textless{}none\textgreater{} &      & 180304 &        &                      \\
explicit\_DM        &  1 & 180326 &   23.88 & $1.025 \times 10^{-6}$ *** \\
position\_in\_doc              &  1 & 180513 &  210.43 & $< 2.2 \times 10^{-16}$ *** \\
discourse\_relation        & 14 & 181200 &  923.02 & $< 2.2 \times 10^{-16}$ *** \\
edu\_depth &  1 & 181551 & 1248.03 & $< 2.2 \times 10^{-16}$ *** \\
\bottomrule
\end{tabular}
\end{table}

Although explicit relation marking is the weakest effect in the table, it is still quite significant, though the strongest effect belongs to discourse nesting depth, closely followed by the relation label. The effect of linear position in the document ranks third based on AIC, and closer to the effect of explicit DMs, given that discourse nesting depth is made available to the model.

\subsection{Does referential structure signal discourse salience?}\label{sec:centering}

\begin{figure}[bth]
\centering
\includegraphics[width=0.9\textwidth]{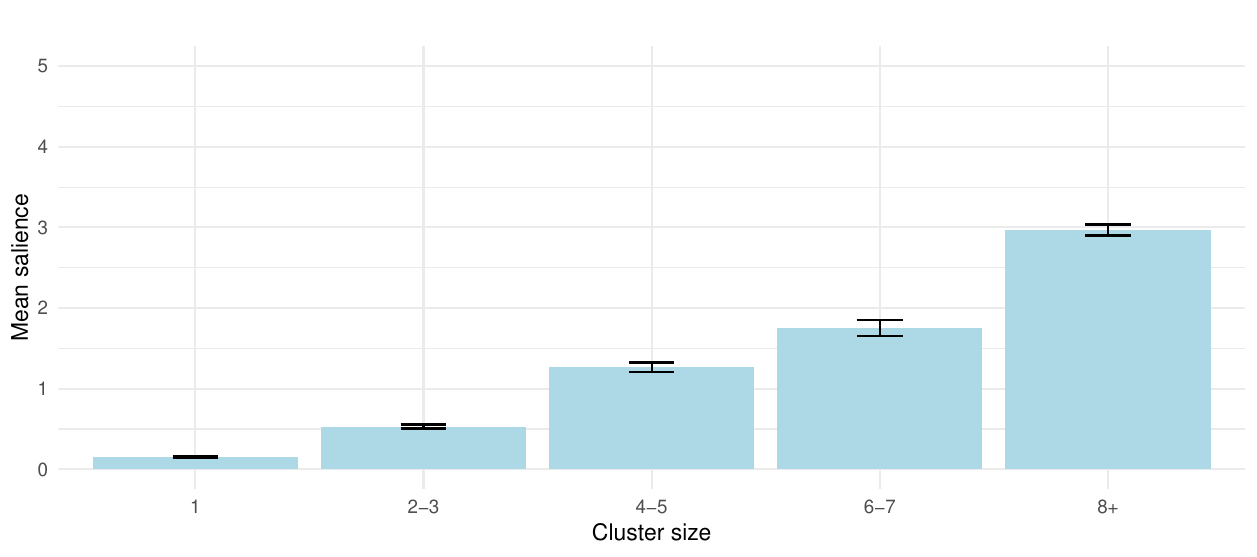}
\caption{Mean salience by cluster size, left to right from singletons with 1 mention to entities with over 8 mentions. All bars differ significantly with GLHT adjusted confidence intervals.}
\label{fig:cluster-size-bins}
\end{figure}

We have seen in Section \ref{sec:rst} that while the position of first mentions in both linear document order and discourse structure plays an important role as a correlate of discourse-level salience, models including a range of properties from all occurrences of an entity are significantly predictive of our salience scores. Rather than taking individual mentions as separate data points, we can also ask whether or how entity-based properties of a cluster of mentions taken together can signal salience. Most trivially, we may expect that entities mentioned \textit{many times} in a document will be more salient than others (cf.~\citealt[184]{Chafe1994}). This is easy to confirm using coreference annotations (see \citealt{Zeldes2022} for coreference criteria in GUM): cluster size correlates strongly with salience scores, with  $r^2=0.4505, p<0.00001$. Figure \ref{fig:cluster-size-bins} gives the average salience for several cluster size ranges along with GLHT-adjusted 5\% confidence intervals. All pair-wise differences are significant.

Singletons, or entities with only one mention, have a mean salience of only around 0.13, while entities with 8 or more mentions have mean scores closer to 3, a very substantial disparity. A confounding factor here is document length, since shorter documents by nature are less likely to have mention chains with very many members. An alternative measure of prevalence for a cluster in a document is considering its dispersion, i.e.~the extent to which an entity has mentions all across the document or just in one or two regions. Work on coreference resolution has shown that the number of entities that are `active' in a given part of a document is usually small, and those with the broadest dispersion tend the be the most important \cite{toshniwal-etal-2020-learning}. Dispersion can be operationalized in several ways
\citep{gries2008dispersions}, but here we follow \citet{Gries2021} and compute scaled Kullback-Leibler Divergence for entity occurrences in each document across 10 partitions (the deciles of the document's token positions), denoted $D_{KL}$. This turns out to be an even stronger correlate of salience than cluster size: $r^2=-0.5352, p<0.00001$, possibly since it can account for salient entities with smaller clusters in shorter documents, which still span multiple portions of the text.

However, neither of these metrics considers the pattern in which repeated mentions occur. For example, Centering Theory identifies Cbs with the anaphoric element of each sentence corresponding to the highest ranked member of Cf in the previous sentence, and predicts that if any element is realized as a pronoun, Cb must be a pronoun. Since Cbs are guaranteed to have been top members of Cf in the previous sentence, and givenness and pronominality are themselves criteria in the Cf ordering (at least in some versions, \citealt{strube-hahn-1999-functional}, \citealt[319]{poesio-etal-2004-centering}), we might expect that Cbs will be locally salient, and that the most prominent entities at the utterance level would be those with pronominal mentions. Centering also suggests that discourse is easier to process when successive utterances have a unique discourse topic, which may lead us to suspect that top-ranked entities in consecutive sentences (i.e.~in a continuation topic environment) may be especially prominent in the local context. A simple proxy to test whether such mentions are also salient at the discourse level would be to use the percentage of pronominal mentions per cluster; however in practice, only about 15\% of entities are ever mentioned as pronouns, meaning this metric would be useless for distinguishing salient entities among the remaining 85\%. Testing its correlation with salience yields a significant but much weaker correlation of $r^2=0.1$. 

Instead, we can simply use the Centering annotations available in the corpus, considering three potential feature types: 1. features based on Cf score; 2. based on proportion of mentions that were the Cb; and 3. sentence transition types. For Cf scores, one challenge is that every sentence contains different amounts of mentions, meaning that a Cf ranking of 2 is more impressive in a sentence with 10 entities than it is in a sentence with just 2. We therefore transform the Cf score of each mention into a percentile based on the total number of entities in each sentence, where Cf rank 2 out of 10 entities means 0.2, and then average the scores across all mentions of each entity to create an entity level variable, whose correlation with discourse-level salience we can test. For the sentence transitions, Centering Theory defines as many as 7 types, which can be ranked by expected coherence as follows \citep{poesio-etal-2004-centering}, starting with the most prominent type, `Continuation':

\vspace{6pt}
\begin{enumerate}
    \item \textbf{Continuation}: Cb is the same as in the previous sentence and ranks highest in the current Cf list.
    \item \textbf{Retention}: Cb is the same but not highest-ranked in the Cf list.
    \item \textbf{Smooth Shift}: Cb changes and the new Cb is highest-ranked in the Cf list.
    \item \textbf{Rough Shift}: Cb changes and new Cb is not highest-ranked in the Cf list.
    \item \textbf{Establishment}: No prior Cb, but the current utterance introduces one.
    \item \textbf{Null}: No Cb in the current utterance despite having one in the previous.
    \item \textbf{Zero}: No Cb in either the previous or the current utterance.
\end{enumerate}

Although this ordinal ranking is not a true ratio scale, for simplicity we test two numeric operationalizations: 
minimum (if an entity is ever in a continuation, its score is 1) and mean rank (treating transitions as a ratio scale, with fractional means being possible). Testing the correlations for our operationalizations we obtain the following ranking sorted by absolute $r^2$ (all significant for $\alpha<0.00001$): 

\vspace{6pt}
\begin{enumerate}
    \item Mean transition rank: -0.12
    \item Mean Cf percentile rank: -0.35
    \item Minimum transition rank: -0.43
    \item Cb proportion: 0.52
\end{enumerate}
\vspace{-4pt}

We observe two main take-aways from this ranking: 1. that presence in a sentence that is a focal point of cohesion (low minimum transition) is a much better predictor of salience than an average across sentences; and 2. being likely to be mentioned again based on Cf is inferior to observing actual repeated mention as the highest ranked anaphoric item in a subsequent sentence based on Centering Theory.

The first finding is perhaps unsurprising if we consider that salient entities can easily appear both in sentences that are cohesive focal points and in other places, but the non-salient entities that appear only in such non-cohesive points can be filtered out by this metric. The second finding to some extent simply echoes the importance of repeated mention, since being Cb at any point guarantees a cluster size >1. Taking all of the Centering metrics together and adding cluster size percentile as a control, we can obtain the model in Table \ref{tab:centering-model}.

\begin{table}[htb]
\centering
\caption{ANOVA table for model: \texttt{salience \textasciitilde ~mean\_cf + cb\_proportion + mean\_transition + min\_transition + cluster\_size}, with predictors sorted by change in AIC based on single term deletions, from the least to the most informative.}
\label{tab:centering-model}
\begin{tabular}{lrrrrr}
\toprule
Term & Df & AIC & LRT & Pr($>$Chi) \\
\midrule
cb\_proportion    & 1 & 47707 &    0.56  & $0.455$ ns~~ \\
\textless{}none\textgreater{} &      & 47708 &        &                      \\
mean\_cf          & 1 & 47723 &   17.14  & $3.471 \times 10^{-5}$ *** \\
mean\_transition  & 1 & 48725 & 1019.23  & $< 2.2 \times 10^{-16}$ *** \\
cluster\_size  & 1 & 48831 & 1124.55  & $< 2.2 \times 10^{-16}$ *** \\
min\_transition   & 1 & 48965 & 1258.46  & $< 2.2 \times 10^{-16}$ *** \\
\bottomrule
\end{tabular}
\end{table}

As the model shows, all Centering predictors remain highly significant, except for Cb proportion. This confirms the suspicion about the second finding above (Cb is collinear with and inferior to cluster size, adding no significant information), but supports an independent effect for sentence transition type, including separate effects for the mean and most cohesive transition value in the cluster.


\vspace{-12pt}
\section{Multilayer analyses}\label{sec:multilayer}
\subsection{Multifactorial models}\label{sec:multifactorial}

To better understand how the variables we have surveyed above all interact, in this section we construct and analyze two kinds of models: 1. descriptive beta-biomial regression models designed primarily to test the significance of each predictor next to competitors in their ability to explain variance in salience scores taken as a ratio-scaled or binary variable; and 2. a predictive classification tree ensemble to evaluate and analyze the relative contribution of each variable for predicting unseen data, both in-domain (the test set) and out-of-domain (the second OOD test set from GENTLE), with all possible feature interactions being considered.

To construct the first type of model we take all of the significant univariate features analyzed above, opting for the strongest representation where multiple ones have been considered. In order to ensure model convergence, all values of categorical variables with fewer than 300 attestations are collapsed into `other' (for example rare dependency labels), and all numerical variables are z-scaled. We also include an additional categorical variable for genre, which will be discussed in detail in the next section. The document from which each observation is taken is treated as a random effect.
Table \ref{tab:mixed-effects-model} gives p-values and AIC scores for single term deletions of the complete mixed effects regression model, with fixed effects sorted by AIC (strongest effects are at the bottom).

\begin{table}[ht]
\centering
\caption{ANOVA Table for full mixed effects model (sorted by AIC)}
\label{tab:mixed-effects-model}
\resizebox{\textwidth}{!}{
\begin{tabular}{lrrrrr}
\toprule
\textbf{Random Effects:} & & & & \\
\midrule
\textbf{Groups} & \textbf{Name} & \textbf{Variance} & \textbf{Std.Dev.} & \\
\midrule
docname & (Intercept) & 0.07264 & 0.2695 & \\
Residual & & 1.54472 & 1.2429 & \\
\multicolumn{5}{c}{Number of obs: 53402, groups: docname, 191} \\
\midrule
\textbf{Fixed Effects:} & & & & \\
\midrule
\textbf{Term} & \textbf{Df} & \textbf{AIC} & \textbf{LRT} & \textbf{Pr($>$Chi)} \\
\midrule
\textless{}none\textgreater{}      &     & 175362 &        &                        \\
position\_in\_sent (morphosyntax)       &  1 & 175364 &   3.78 & $0.05177$ . \\
explicit\_dm\_proportion (eRST)     &  1 & 175383 &  23.11 & $1.528 \times 10^{-6}$ *** \\
mean\_cf (Centering)                &  1 & 175398 &  38.60 & $5.214 \times 10^{-10}$ *** \\
genre                    & 15 & 175411 &  79.41 & $8.958 \times 10^{-11}$ *** \\
cf (Centering)                       &  1 & 175457 &  97.07 & $< 2.2 \times 10^{-16}$ *** \\
number (morphosyntax)                   &  1 & 175553 & 192.75 & $< 2.2 \times 10^{-16}$ *** \\
dependency\_relation (morphosyntax)     & 17 & 175581 & 252.93 & $< 2.2 \times 10^{-16}$ *** \\
position\_in\_doc       &  1 & 175664 & 304.05 & $< 2.2 \times 10^{-16}$ *** \\
edu\_depth (eRST)  &  1 & 175737 & 376.75 & $< 2.2 \times 10^{-16}$ *** \\
discourse\_relation (eRST)     & 30 & 175811 & 508.72 & $< 2.2 \times 10^{-16}$ *** \\
entity\_type             &  9 & 175898 & 554.20 & $< 2.2 \times 10^{-16}$ *** \\
part-of-speech (morphosyntax)          &  8 & 175917 & 570.72 & $< 2.2 \times 10^{-16}$ *** \\
mean\_transition (Centering)        &  1 & 176147 & 787.44 & $< 2.2 \times 10^{-16}$ *** \\
min\_transition (Centering)         &  1 & 176645 & 1285.59 & $< 2.2 \times 10^{-16}$ *** \\
cluster\_size &  1 & 176755 & 1394.77 & $< 2.2 \times 10^{-16}$ *** \\
cluster\_divergence      &  1 & 177936 & 2576.25 & $< 2.2 \times 10^{-16}$ *** \\
\bottomrule
\end{tabular}
}
\end{table}

All fixed effects are significant except position in sentence, which is marginal at $p=0.05177$. The model has a good fit to the data, with conditional $R^2_c=0.594$ (variance explained by all effects) and marginal $R^2_m=0.575$ (fixed effects only). It also achieves a root mean squared error (RMSE) score of 1.229, meaning its predictions are off by that amount on average. By comparison, the baseline of predicting the mean salience value every time would score 1.948, off by almost two salience scores. In binary classification, predicting whether salience will be above 0, the same model scores 69.8\% accuracy, over a majority baseline of 63.49\% (predicting no mention is salient), a modest but significant outcome. We can also confirm that the model is not overfitted by testing it on the held-out test partition, where RMSE is 1.194 improving on a baseline of 1.886 (lower is better) and binary accuracy is 70.22\% over a baseline of 62.37\%, suggesting the test set is slightly easier for the model to handle, though it still surpasses the baseline by only $\sim$8\%.

\begin{figure}[tbh]
\centering
\includegraphics[width=0.85\textwidth, trim={0cm 0.8cm 0cm 2.4cm}, clip]{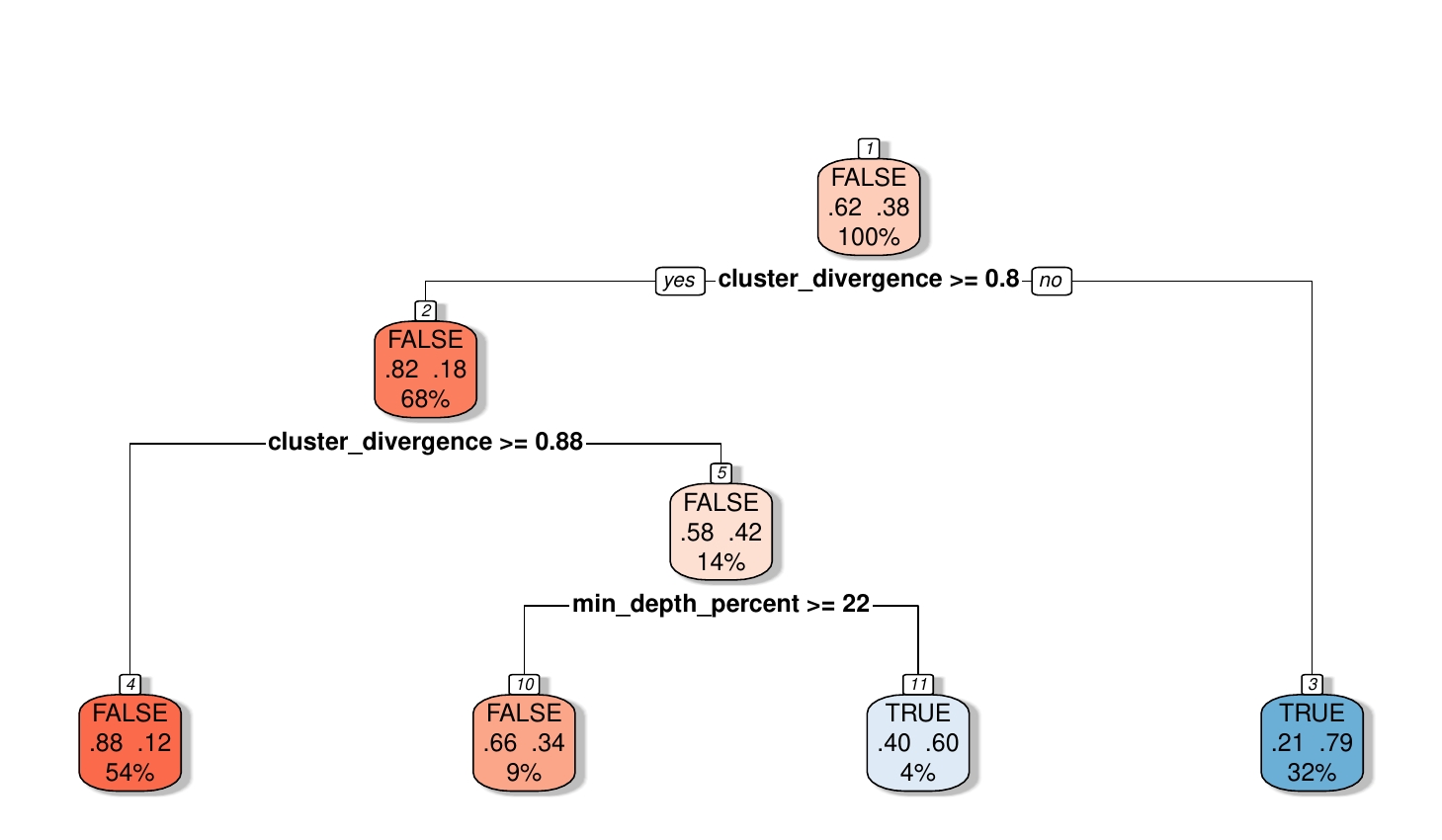}
\caption{A single decision tree with four bins. Blue bins correspond to predictions of salience = \textsc{true}, and each bin indicates the percentage of data it includes and the proportion of \textsc{false} versus \textsc{true} after application of the rule that defines it.}
\vspace{-12pt}
\label{fig:decision-tree}
\end{figure}

However, the purpose of the linear model, which is rather simple and includes no interactions between variables, is not to maximize prediction accuracy -- it is merely meant to confirm the significance of all predictors when used in tandem. To produce a more accurate model, which we can then analyze to understand the relative importance of each variable and how they combine, we construct a Random Forest ensemble model using the Extra Trees algorithm \citep{GeurtsErnstWehenkel2006}, which offers a simple, but powerful and interpretable model accounting for variable interactions. We focus on the classification setting, i.e.~predicting whether salience will be >0, meaning an entity is summary-worthy for at least one summary. As tree ensembles, Random Forest models rely on multiple hierarchical decision trees, in which variables can be used to split the data based on value thresholds into bins, which in our case will be more or less likely to contain salient mentions. To illustrate this, Figure \ref{fig:decision-tree} plots one such shallow tree, limited to producing 4 bins.

The figure shows that the algorithm first selects the cluster-divergence variable, setting the threshold at 0.8. Recall that our normalized divergence is bounded between 0 and 1, with 0 corresponding to perfect homogeneity across the document, and 1 to appearing in just one decile of the document, meaning that not exceeding 0.8 divergence predicts a relatively dispersed entity, therefore leading to a prediction of \textsc{true} (i.e.~salient) in the rightmost bin. The algorithm then splits again based on divergence on the left, this time with a higher threshold of 0.88, and for observations failing that test, next opts to split based on the minimum discourse depth achieved by the entity, i.e.~how close any of its mentions are to the discourse root (see Section \ref{sec:rst}). In this case if the entity is in the bottom 22\% of discourse depth, the prediction is for salience=\textsc{true}, since at this node, 60\% of mentions would correspond to salient entities (though this only applies to 4\% of the data, as indicated in the second bottom node from the right). The entire tree ensemble consists of many such trees (in our case, the default 100) voting together on the predicted outcome (salient or not). Each tree is exposed to a different subset of observations and variables to improve stability and reduce overfitting.

\begin{figure}[htb]
\centering
\includegraphics[width=1\textwidth]{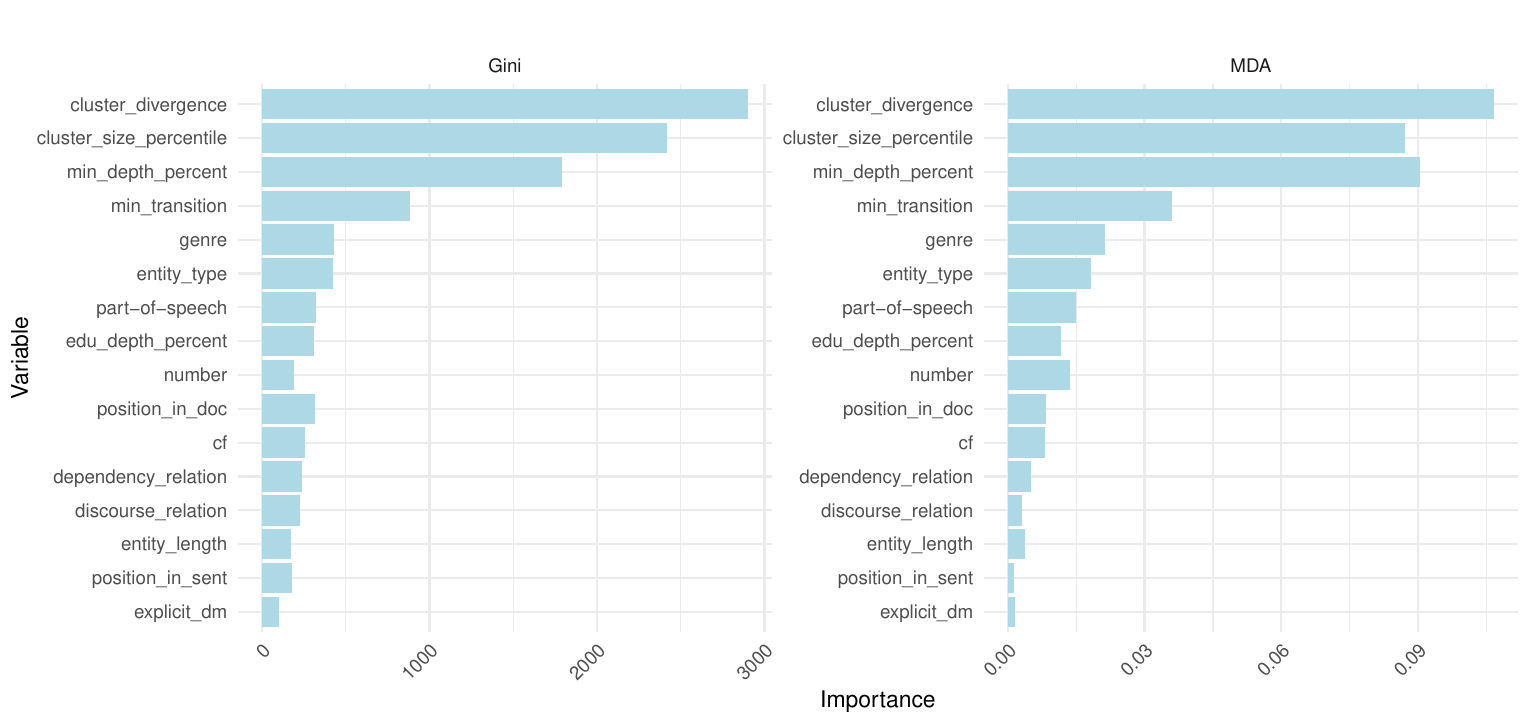}
\caption{Variable importances for the Random Forest model using Gini impurity reduction (increase in node purity from using a variable for a split) and permuted Mean Decrease Accuracy (MDA, how much worse accuracy would be if this variable is perturbed).}
\label{fig:variable-importances}
\vspace{-12pt}
\end{figure}

With this type of decision tree in mind, we analyze the impact of variables using two different metrics: Gini indices, corresponding to the relative increase in node purity achieved by using a variable for a split, and mean decrease in accuracy (MDA), indicating, on average, how much worse the prediction accuracy of each tree in an ensemble would become if the values of that variable were randomized during testing. A high Gini index corresponds to highly imbalanced nodes (for example the split on cluster-divergence >= 0.88 in Figure \ref{fig:decision-tree} leads to a fairly impure node with a 58-42 split on the right, but a purer node with a more lopsided 88-12 split on the left). Gini indices are known to be vulnerable to overestimating the importance of variables which can easily be used multiple times in a tree (for example numerical, but not binary variables), while MDA tends to underestimate collinear variables, since randomizing just one of two very similar variables may have limited impact, although having at least one of them may be essential. We therefore explore both metrics and emphasize that numbers are mainly meaningful for internally ranking variables against each other. Figure \ref{fig:variable-importances} provides variable importances using Gini and MDA for all the variables in our model. The variables are sorted from most to least important, top to bottom, based on the mean z-score for each variable across both metrics.

Both metrics agree in placing cluster KL divergence (i.e.~dispersion) at the top, and cluster size percentile and minimum discourse tree depth second and third, albeit in different orders. The Centering Theory minimum transition type comes next, followed by weaker predictors, led by genre, entity type (including \textsc{person} vs. other types), part-of-speech (including pronominalization), and the discourse nesting depth at the instance level. Grammatical function (including subjecthood) ranks relatively low at fifth from the bottom on average across scaled metrics, while explicit discourse markers and position in the sentence are close to negligible, given the other variables.

Since Random Forests are more prone to memorizing training data than linear models, we evaluate the model's performance only on unseen data. In terms of accuracy, the model achieves 83.6\% correct predictions for mention salience over a baseline of 63.4\% in the test set. We can also test the model on the out-of-domain GENTLE test set, with 8 unseen genres -- in this case, the model cannot make effective use of the genre variable.
We observe a degradation to an accuracy of 80.9\%, less than 3\% below results for the in-domain test set. At the same time we observe that the baseline accuracy of always selecting `not salient' for GENTLE is also lower at 60.2\%, meaning the model performs close to comparably out-of-domain.

These results appear to be somewhat contradictory: on the one hand, genre was ranked fifth most important by the ensemble, expected to lead to over 3\% degradation, but on the other, testing out-of-domain with no relevant genre information leads to little degradation over the baseline for that dataset. We suspect that this is because the other top variables do a better predictive job in the GENTLE genres: for example, the correlation between salience and divergence is $r=-0.71$ for the training set, but $r=-0.78$ for GENTLE, and with cluster size percent it is $r=0.685$ compared to $r=0.725$. This means the top features may make the GENTLE set easier than expected, compensating for the lack of genre information. But if genre is not so important, why does it still rank above established correlates of salience such as entity type, part of speech, grammatical functions and linear position? We turn to this question next using genre-based error analysis.
\vspace{-9pt}

\subsection{Genre and communicative intent}\label{sec:genre}

In the sections above, we have mostly looked at correlates of salience as general properties, which apply regardless of the nature of the text in question: being the subject increases the likelihood of a higher salience score, as does being a \textsc{person} entity, being in an \textsc{organization} relation (i.e.~especially being mentioned in a heading), etc. However we already observed in Section \ref{sec:people} that entity type biases vary by genre (on average \textsc{place} entities are more salient than people in travel guides), and some categories are much rarer or even absent in specific types of language -- for example conversations simply do not contain headings. It is also well known that genre and modality can interact with choice of referring expression types, perceptual salience and priming effects \citep{Toole1996effect,Travis_2007,VinelEtAl2021}. To understand why genre can have a strong effect on models' ability to predict salience, consider example \ref{ex:place-not-person2}, repeated here as \ref{ex:place-not-person2-rep}.

\ex. [\textit{Oakland}]$_{5}^{place}$\textit{nurtured novelists} [\textit{Amy\,Tan}]$_{0}^{person}$\textit{and}\,[\textit{Maya\,Angelou}]$_{0}^{person}$\label{ex:place-not-person2-rep}

This sentence appeared in a travel guide, and unsurprisingly, Oakland is its most salient entity, not Amy Tan or Maya Angelou. However if this were for example an essay (perhaps comparing Maya Angelou and Amy Tan's writing), or a biography of either one of them, then they would most likely be ranked higher, and Oakland might not be predicted to be salient, despite appearing as the subject in the sentence. In other words, genres create prior expectations for the kinds of entities that would be salient (e.g.~in biographies: a single, repeatedly mentioned, well-dispersed, proper named, grammatically singular, \textsc{person} entity).

To explore how genre figures into the web of variables that signal salience, we can exploit our Random Forest model's permutation importance evaluation (MDA in Figure \ref{fig:variable-importances} above) and look specifically at the scenario in which the values of the genre variable are randomized to assess its importance. In this scenario, all other variables in the data have their true values, but genre has been shuffled, meaning it can cause the model to get predictions wrong that it would have gotten right if genre had been encoded correctly. We extract the top positive and negative predictions by probability assigned in the shuffled scenario, for which the model errs on shuffled data but succeeds on the original data, ranked by the difference in probability of the two scenarios (i.e.~in the top positive instance, the shuffled scenario might receive a probability of 0\% while the normal data might lead to 100\% certainty for a salient judgment). The model is trained on the train set and for the data to be analyzed we use the development partition. The top 5 positive and 3 out of 5 negative cases all concern pronouns, as illustrated by \ref{ex:shuff-fiction}--\ref{ex:shuff-essay}:

\ex. \textit{I saw that my father’s face was shining with pride, and} [\textit{his}]$_{5}^{fiction}$ \textit{bearing had in it a new hauteur} (p(orig)=0.878, p(shuff)=0.496)\label{ex:shuff-fiction}

\ex. \textit{You're so stupid thinking} [\textit{I}]$_{4}^{conversation}$ \textit{spent the night} \\(p(orig)=0.768, p(shuff)=0.394)\label{ex:shuff-conversation}

\ex.\textit{If you hear a lot of awkward pauses or “ah”s or “um”s,} [\textit{your}]$_{0}^{wikihow}$ \textit{joke isn’t ready} (p(orig)=0.489, p(shuff)=0.876)\label{ex:shuff-whow}

\ex. [\textit{This}]$_{0}^{essay}$ \textit{is a trend that bears more scrutiny than it has received} (p(orig)=0.471, p(shuff)=0.604)\label{ex:shuff-essay}

These examples show that genre matters a great deal in classifying the likelihood that pronouns will be discourse salient: speakers in a face-to-face conversation, appearing as `you' and `I' in the dialog, are very likely to be mentioned in a summary, but this is less true in other genres, such as news, in which some quoted participants may be tangential or appear only briefly, or how-to guides, in which people are not typically the focus. Human characters are also very likely to be salient in fiction, even in the third person, as in \ref{ex:shuff-fiction}. By contrast, the generic `you' in a how-to guide in \ref{ex:shuff-whow} is a very non-salient entity, but this is hard to identify based on our features without knowing the genre. Similarly a demonstrative `this' can be salient in a situated deictic context, but in the context of an essay as in \ref{ex:shuff-essay}, it refers to an abstract notion that is not central.

For non-pronominal mentions we can also see why genre may matter: 

\ex. [\textit{Wikinews}]$_{2}^{interview}$ \textit{interviews meteorological experts on Cyclone Phalin } (p(orig)=0.647, p(shuff)=0.438)\label{ex:shuff-interview}

\ex. \textit{I made noises with} [\textit{my heels}]$_{0}^{fiction}$ \textit{but they were too loud so I stopped.} (p(orig)=0.326, p(shuff)=0.531)\label{ex:shuff-fiction2}

In the shuffled false negative in \ref{ex:shuff-interview}, Wikinews is an organization which is never mentioned again after the heading of the interview, making it seem non-salient, yet because it occupies the salient `interviewer' slot in the interview genre, two summaries mention it (salience=2). In other words, when summarizing an interview, part of the communicative intent to be conveyed often includes mentioning `who interviewed whom', and a model that knows about genre has seen this pattern before; but without genre information, the prediction fails. In \ref{ex:shuff-fiction2}, the original model anticipates that the oblique inanimate object `my heels' will not be salient, but the model with shuffled genre information has a false positive. When could such a two-word, plural inanimate object be salient to a document's communicative intent? If we consider contexts like how-to guides, in which footwear can be salient, it is easy to find similar salient examples, as in \ref{ex:slippers} from a guide to ballet dancing:

\ex. \textit{Always wear} [\textit{ballet slippers}]$_{1}^{wikihow}$\label{ex:slippers}

In sum, it seems that salient entities can belong to different prototypes or profiles, and while some are usually likely across genres (well dispersed human entities with many mentions serving as a pronominal subject), others are more specialized, such as `you' in a conversation (salient) versus in a how-to guide (usually not). If prototypes or profiles are a good way of thinking of salient entities across text types, we would like to know whether such profiles can be extracted from the data using quantitative methods. We turn to this question in the final section before our closing discussion.
\vspace{-9pt}

\subsection{What kinds of salient entities are there?}\label{sec:clustering}

In order to study the kinds of entity profiles found in our data, we need a metric to indicate which observations should be considered similar to each other, which can then be used to cluster similar cases into larger groups. Since our data is large and relatively high-dimensional, dimensionality reduction techniques provide a good way of summarizing the data and allowing for interpretable visualization. 

\begin{figure}[t!bh]
\centering
\includegraphics[width=0.95\textwidth]{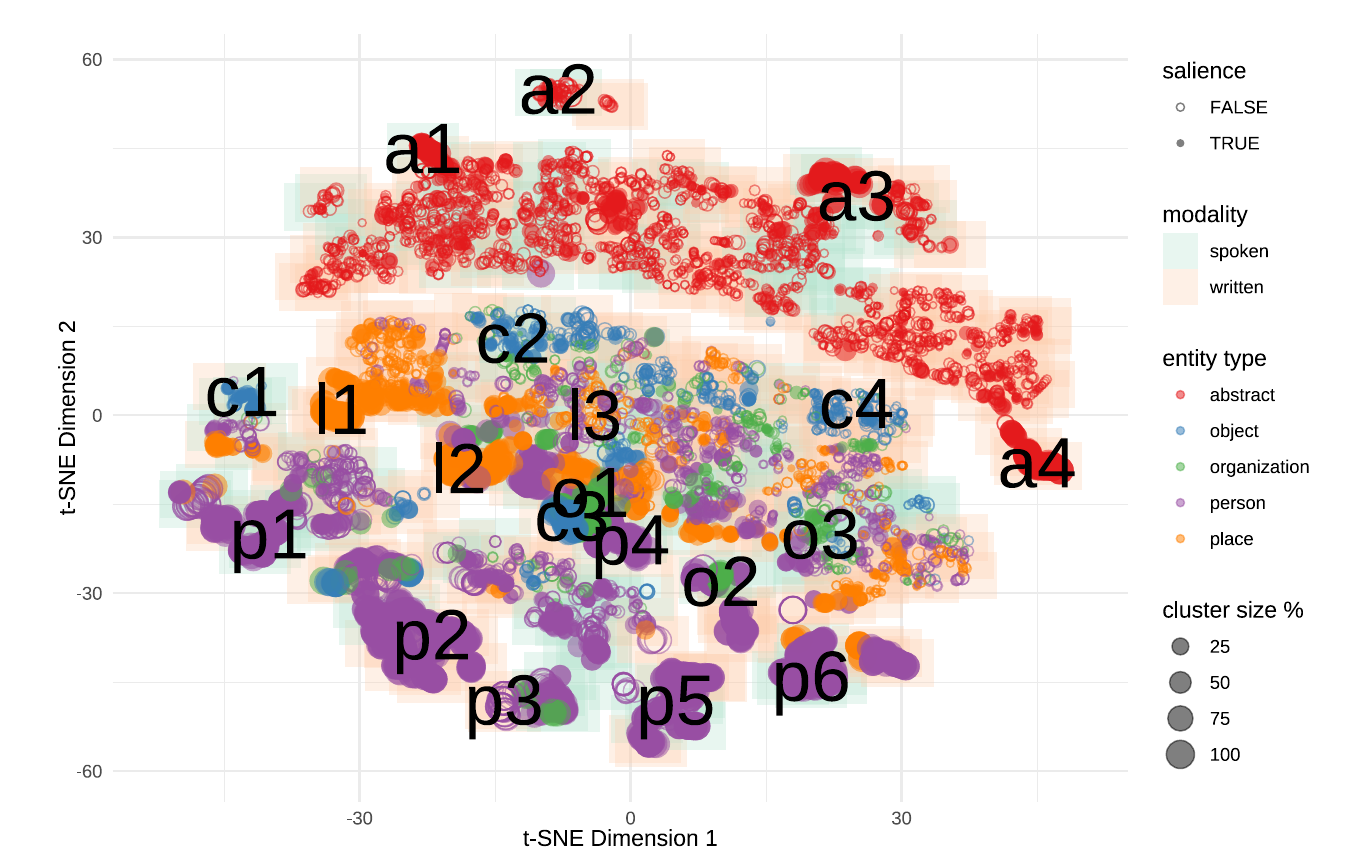}
\caption{A t-SNE plot of mentions with five entity types in the dev set data.}
\label{fig:tsne}
\end{figure}

Figure \ref{fig:tsne} shows a t-SNE plot of the development set data, where the x and y-axes are reduced dimensions selected by the algorithm to represent proximity. To make the plot more tractable, we focus on just the five most common entity types, which are distinguished by color: \textsc{person} (purple, with some regions manually labeled with `p' and a number), \textsc{abstract} (red, regions of interest labeled with `a'), \textsc{organization} (green, labeled `o'), concrete \textsc{objects} (blue, `c') and \textsc{place} or location entities (orange, `l'). Hollow points indicate non-salient mentions, while filled ones are salient, and circle size indicates the percentile size of the entity cluster in mentions (i.e.~an entity with the most mentions in a document has the largest circle, corresponding to 100\% of the maximum available cluster size). Background shading tiles behind the points indicate the genre's modality, with regions of predominantly spoken data shaded in light green, while more written data is shaded a light orange.

Looking at the plot we can recognize several large clusters distinguished by entity types, with purple \textsc{person} entities clustering at the bottom in several sub-regions, and \textsc{abstract} entities, which have much smaller clusters on average and less salient mentions, occupying a crescent at the top. Other entity types are more mixed in the middle of the plot, but distinct sub-clusters emerge, which we can inspect below using the examples nearest to the centroid of each labeled area.

\vspace{12pt}
\noindent\textbf{Person entities: p1--p6}\quad Most \textsc{person} entities are salient and correspond to large clusters, especially in regions p1, p2, p5 and p6. Looking at examples from these regions, it seems clear that this part of the t-SNE space skews to generally favor pronouns, but of somewhat different kinds. Region p1 comes primarily from spoken data and includes somewhat salient third person entities being discussed as subjects, as in \ref{ex:bezos-he}, though the region's overall mean salience is only 2.26. Region p2 also includes pronouns, but primarily ones corresponding to salient main characters in written fiction, with some of the largest cluster sizes and broadest dispersion values \ref{ex:protagonist1}, and a uniform salience score of 5. Region p3 is non-salient ($\oslash=0.89$) and seems to capture generic uses of `you', as in \ref{ex:generic-you}, while p5 and p6 focus on written and spoken first person `I' respectively, with mean scores of 3.78 for the former and a uniform 5 for the latter. Chains with higher concentrations of nominal mentions are exemplified by p4 ($\oslash=4.72$), which contains sometimes longer, less often mentioned but named entities, as in \ref{ex:person-nominal1}--\ref{ex:person-nominal2}.

\ex. \textit{everyone in my neighborhood pays a higher percentage of their income in taxes than} [\textit{he}]$_{5}$ \textit{pays} (podcast about Jeff Bezos)\label{ex:bezos-he}

\ex. [\textit{She}]$_{5}$ \textit{waves, antlike, then crawls through a dormer window}\label{ex:protagonist1} (fiction)

\ex. [\textit{He}]$_{5}$ \textit{watched the second hand trotting round the dial of his watch} \label{ex:protagonist2} (fiction)

\ex. [\textit{You}]$_{0}$\textit{’re great at sawing and hammering, but architecture is not your forte} (hypothetical, from an essay)\label{ex:generic-you}

\ex. [\textit{Sarvis}]$_{5}$ \textit{has garnered double digits in opinion polls} (political interview)\label{ex:person-nominal1}

\ex. \textit{Wikinews interviews} [\textit{Christopher Hil}]$_{5}$\textit{, U.S. Republican Party presidential candidate} (political interview)\label{ex:person-nominal2}

\noindent\textbf{Abstract entities: a1--a4}\quad While the vast majority of \textsc{abstract} mentions are non-salient singletons (small hollow circles across the top of the plot in red), there are four focal points of non-singleton mentions: three of these are highly salient -- a1, a3 and a4 -- with especially a3 and a4 having rather large, salient clusters sizes (salience $\oslash=3$ and $4.71$). The latter two regions correspond respectively to pronominal mentions of abstract entities under discussion \ref{ex:a3-1}--\ref{ex:a3-2} and short nominal ones, often recurring technical terms in academic writing which are not pronominalized \ref{ex:a4-1}--\ref{ex:a4-2}. 

\ex. \textit{while} [\textit{it}]$_{5}$ \textit{didn't find it on each and every specific count, that was solely because the instruction wasn't given even though requested} (courtroom transcript, referring to a pivotal jury instruction)\label{ex:a3-1}

\ex. [\textit{Its}]$_{5}$ \textit{roots are deep in our core psychological and biological being, and it is one of our most intimate feelings} (from an essay, talking about `fear')\label{ex:a3-2}

\ex. \textit{measures of perceived} [\textit{discrimination}]$_{5}$ \textit{in a large American sample} (in an academic article about discrimination)\label{ex:a4-1}

\ex. \textit{The arrow on the map points to the} [\textit{Eegimaa}]$_{5}$ \textit{speaking area}  (academic article about Eegima language)\label{ex:a4-2}

As the examples show, mentions in a3 tend to have the more salient grammatical functions (subject, possessor), while a4 more typically has prepositional and compound modifiers.

Region a1 is similar to a4, but has longer phrases with somewhat fewer mentions and lower scores ($\oslash=2.76$), as shown in \ref{ex:a1-1}--\ref{ex:a1-2}, while the isolated region a2 is least salient ($\oslash=0.41$), with mentions from non-singleton chains that are nevertheless not salient. These are often cases of discourse deixis \citep{cornish2007demonstratives} as in \ref{ex:a2-1}--\ref{ex:a2-2}, where the discourse picks up a non-nominal antecedent for a brief discussion which is not a main theme of the text.

\ex. \textit{Here are the postulates of }[\textit{Dalton’s atomic theory}]$_{5}$ (chemistry textbook)\label{ex:a1-1}

\ex. \textit{our public health service is working hard to .. limit the risk of any spread of} [\textit{the disease}]$_{5}$ (political speech about Covid in New Zealand)\label{ex:a1-2}

\ex. \textit{but something in the way he says} [\textit{it}]$_{0}$\textit{ .. suggests that there simply are no other options for him} (Reddit discussion about a movie scene)\label{ex:a2-1}

\ex. [\textit{This}]$_{0}$\textit{.. reminds readers and listeners of the ProPublica story that exposed how little he and the billionaire class pay in taxes} (podcast about Bezos)\label{ex:a2-2}

\noindent\textbf{Location entities: l1--l3}\quad There are only two large clusters of frequently recurring \textsc{place} entities, l1 and l2 with mean salience of 3.79 and 4.58 respectively. The l1 region is populated mainly by non-named mentions of places talked about in travel guides, either as predicate phrases (what a place is or isn't) or in other definite non-named mentions like `the city', `the island', `the country' etc. \ref{ex:l1-1}--\ref{ex:l1-2}. By contrast, l2 is more specialized on named places, which most often occur in prepositional phrases \ref{ex:l2-1}--\ref{ex:l2-2}.

\ex. \textit{Neiafu town is} [\textit{the centre of activity}]$_{5}$ (travel guide to Vava'u)\label{ex:l1-1}

\ex. \textit{Uptown also has some of the best in} [\textit{the city's}]$_{5}$ \textit{vintage architecture} (guide to Oakland)\label{ex:l1-2}

\ex. \textit{These restrictions have played a key role in keeping COVID-19 out of }[\textit{New Zealand}]$_{5}$ (speech about Covid in New Zealand)\label{ex:l2-1}

\ex. \textit{free Black people from} [\textit{the North}]$_{5}$ \textit{enlisted} (textbook about the Civil War)\label{ex:l2-2}

Region l3 contains typical examples of non-salient, often indefinite common-noun places ($\oslash=0.11$) -- the two closest examples to the centroid happen to come from Reddit, but the region has a range of genres, spoken and written:

\ex. \textit{Proposal happens in} [\textit{an Escape Room}]$_{0}$\textit{, there's no longer a game }(Reddit)

\ex. \textit{What kind of} [\textit{environment}]$_{0}$ \textit{could "The Wild" be?} (Reddit)

\noindent\textbf{Concrete objects: c1--c4}\quad Concrete objects are rarely salient and typically have smaller mention clusters. The only region with truly high salience is c3 ($\oslash=4$), with the next most salient c1 averaging at just 1.81. The salient mentions in c3 appear to be subject and object nominals, with well-dispersed spreads despite small clusters, and which play an important role in their documents, as shown in \ref{ex:c3-1}--\ref{ex:c3-2}. The salient mentions in c1 tend to be pronominal subjects and objects \ref{ex:c1-1}, while the least salient regions c2/c4 are mainly populated by non-named, longer, multi-word singletons, such as `the shelf by Steven’s bed' (fiction) or `this part on the left foot' (vlog, talking about foot pain).

\ex. [\textit{Sensitive government document}]$_{3}$ \textit{found on rainy Ottawa street} (news)\label{ex:c3-1}

\ex. \textit{Isabelle does not want to sell} [\textit{the letters}]$_{5}$ \textit{unless we are forced to do so by the judge} (letter discussing the fate of the scandalous Warren G. Harding--Carrie Phillips letters)\label{ex:c3-2}

\ex. [\textit{They}]$_{5}$ \textit{are really cute} (referring to shoes a vlogger has just bought)\label{ex:c1-1}

\noindent\textbf{Organizations: o1--o3}\quad Finally looking at \textsc{organization} entities, salient mentions are rare and only appear in 3 clusters. Regions o1 and o2 have the highest scores (4.14 and 5), with o1 showing more named entities \ref{ex:o1-1}--\ref{ex:o1-2} and o2 favoring non-named mentions \ref{ex:o2-1}. Region o3 is non-salient ($\oslash=1.06$) and primarily comes from singletons in spoken data, including courtroom mentions of organizations or official entities which are less pivotal to the gist of the argument, as in \ref{ex:o3-1}.

\ex. \textit{The country is becoming an armed camp, }[\textit{the Government}]$_{4}$ \textit{preparing for civil war} (letter by Nelson Mandela about the South African Government)\label{ex:o1-1}

\ex. [\textit{Environment Canada}]{$_4$} \textit{spokesperson Julie Hahn said they apply to letters of complaint} (news)\label{ex:o1-2}

\ex. \textit{collateral estoppel would prevent retrial without regard for} [\textit{the jury's}]{$_5$} \textit{prior finding} (courtroom transcript)\label{ex:o2-1}

\ex.\textit{at the trial,} [\textit{the State}]{$_{1}$}\textit{, and it can be found in the transcript at pages 774 and 775, admitted as much}\label{ex:o3-1}

These prototypes are by no means an exhaustive list of salient or non-salient entity types, but offer some idea of the recurring patterns found in our data, and the variety we observe in general types versus genre specific patterns.

\vspace{-9pt}

\section{Discussion}

At the outset of this paper we asked how discourse-level salience is expressed for each entity mentioned, and specifically focused on the interactions between overt markings known from previous studies, such as subjecthood, definiteness, entity type, and more. Although the basic correlations between these are a given (subjects, people and definite phrases are all more salient than their alternatives on average), every generalization has exceptions: people rarely matter in travel guides, repeated generic pronouns are not salient in how-to guides, and so on. 

A key distinction between the features we have used to analyze salient entities concerns the difference between entity-based properties of an entire cluster of mentions, and instance-based ones applying at a particular mention. Traditionally important grammatical properties such as subjecthood have weak predictive power because nearly every sentence has a subject in English, but many sentences are relatively unimportant. This observation sheds light on the power of aggregate metrics including obvious entity-level properties such as cluster size or dispersion, and more refined ones such as minimum discourse tree depth achieved for any mention of an entity, or the average or `best' transition in the framework of Centering Theory, all of which turned out to survive in a multifactorial model next to commonly postulated features of salient mentions at the utterance level.

An interesting result from the more complex tree ensemble model is the consistently high ranking of not only entity prevalence metrics such as divergence or cluster size, but also high-level semantic and pragmatic features, such as entity types, discourse relations and Centering features over the more often discussed form features, i.e.~grammatical function, linear ordering (within sentences or the document) or NP form (part of speech, number etc.). Taken together with the other findings above, this suggests that much more attention should be devoted to discourse structure and phenomena above the sentence level in the study of formal linguistic markers of entity salience.

Another interesting consequence of the results is an indirect validation of the relevance of insights from theories such as Centering or (enhanced) Rhetorical Structure Theory to the independently operationalized salience phenomena studied here. Because the summaries we used to obtain salience ratings are separate from the text analyzed using Centering or eRST, the correlation of scores with constructs from those theories and their continuing significance in models containing simpler alternatives, such as frequency, grammatical function or pronominalization, all point to the validity of the relevant theoretical concepts. This finding is non-trivial, as Centering transitions, nuclearity and hierarchical nesting depth, as well as discourse relation types and to some extent even discourse marker signaling turn out to be relevant to what summaries prioritize more or less in a text.

An important methodological point worth discussing in light of this last finding is the value and potential biases of the operationalization chosen in this study. Throughout this paper, we have assumed that summary-worthiness is a good measure of salience at the discourse level, at least when coupled with multiple summaries to alleviate the subjectivity inherent to the summarization task.\footnote{In fact, we are currently experimenting with applying the same method to scoring proposition salience, where this seems even more essential, see \cite{ZeldesEtAl2026}.} Although it is likely that different summaries could change some results in subtle ways, we posit that even if completely new summaries were collected, whether using human or LLMs inputs, the features characterizing salience would remain similar, provided that summarization follows a `general purpose' style. By contrast, and somewhat trivially, aspect-based summarization instructing summarizers to focus for example on people would of course bias results systematically in a different way. We are hopeful that future studies examining salience from different perspectives will engage with the properties studied here and help us to discover the ways in which salience-as-summary-worthiness overlaps with, or differs from, other views of what makes entities important in discourse.

\vspace{-10pt}

\bibliographystyle{plainnat}
\bibliography{article}

\vspace{-16pt}

\section*{Appendix: Glossary}\label{sec:glossary}

\subsection*{Terms}

\begin{itemize}
    \item \textbf{entity-based} - an aggregate property extracted across mentions of an entity, rather than a specific mention (for example total number of mentions)
    \item \textbf{GLHT} - General Linear Hypothesis Testing \cite{HothornEtAl2008}, a technique compensating for compound alpha errors in multiple pairwise significance comparisons of large numbers of groups
    \item \textbf{instance-based} - a property applying to a specific instance or mention of an entity (for example grammatical function in a specific sentence)
    \item \textbf{nucleus} - the discourse unit in an RST/eRST discourse relation to which the relation points, the source of the relation being the satellite
    \item \textbf{POS} - Part of speech (see \citealt{petrov-etal-2012-universal})
    \item \textbf{satellite} - a subordinate discourse unit in RST/eRST which has some directed discourse relation to a more prominent unit, which is its nucleus
    \item \textbf{singleton} - an entity with only one mention in the document
    \item \textbf{UD} - Universal Dependencies - a formalism for cross-linguistic dependency syntax analysis (see \citealt{de-marneffe-etal-2021-universal})

\end{itemize}

\subsection*{UD relation labels and part-of-speech tags}

Lower case labels give dependency relations and upper case labels represent universal part-of-speech tags. Words exemplifying a label within examples in brackets are italicized.

\vspace{6pt}

\begin{itemize}
    \item \textbf{csubj} - UD label for subject clause heads (`That Kim \textit{arrived} was good')
    \item \textbf{dep} - UD label for an unspecified dependency, usually in a non-canonical structure for which no other label exists
    \item \textbf{expl} - UD label for expletive pronouns (`\textit{it} was raining')
    \item \textbf{iobj} - UD label for indirect objects (`Give \textit{Kim} the book')
    \item \textbf{list} - UD label for conjoined orthographic list items such as bullets or subsequent parts of an address
    \item \textbf{nmod:poss} - UD label for possessive pronouns and genitive 's modifiers (`\textit{my}', `\textit{Kim}'s')
    \item \textbf{nsubj} -  UD label for nominal subjects (`\textit{Kim} arrived')
    \item \textbf{obl:unmarked} -  UD label for adverbially used noun heads (`meet next \textit{week}')
    \item \textbf{PRON} - Universal POS tag for pronouns
    \item \textbf{PUNCT} - Universal POS tag for punctuation
    \item \textbf{reparandum} - UD label for the head of a repaired phrase (`The \textit{ma-} -- we saw the man.')
    \item \textbf{vocative} - UD label for vocatives, i.e.~addressees of an utterance being called out explicitly (e.g.~`\textit{Kim}, come here!')

\end{itemize}
\end{document}